\pdfoutput=1

\documentclass[11pt]{article}

\usepackage[]{acl}

\usepackage{times}
\usepackage{latexsym}

\usepackage[T1]{fontenc}
\usepackage{CJKutf8}

\usepackage[utf8]{inputenc}

\usepackage{microtype}

\usepackage{inconsolata}

\usepackage{multirow}
\usepackage[T5]{fontenc}

\usepackage{listings}
\usepackage{tabularx}
\usepackage{ltablex}
\usepackage{longtable}
\usepackage{graphicx}
\usepackage{colortbl}

\usepackage{pythonhighlight}

\usepackage{supertabular}
\usepackage{xltabular}
\usepackage{amsmath}
\usepackage{amssymb}
\usepackage{mathrsfs}
\usepackage{stmaryrd}

\usepackage{adjustbox}
\usepackage{lineno}
\usepackage{array}
\usepackage{booktabs} 

\DeclareRobustCommand{\huggingface}{%
  \begingroup\normalfont
  \vspace{-0.2em}%
  \raisebox{-0.4em}{%
  \includegraphics[height=1.5em]{figures/huggingface_logo.png}%
  }%
  \kern 0.4em%
  \endgroup
}

\DeclareRobustCommand{\github}{%
  \begingroup\normalfont
  \vspace{0.5em}%
  \raisebox{-0.2em}{%
  \includegraphics[height=1.2em]{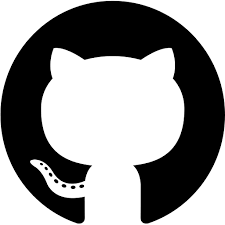}%
  }%
  \kern 0.4em%
  \endgroup
}

\DeclareRobustCommand{\mail}{%
  \begingroup\normalfont
  \vspace{0.em}%
  \raisebox{-0.2em}{%
  \includegraphics[height=1.em]{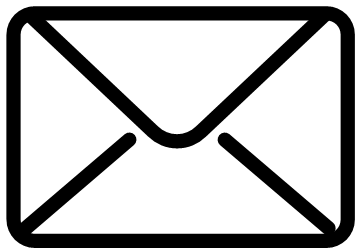}%
  }%
  \kern 0.4em%
  \endgroup
}

\definecolor{custom_light_blue}{rgb}{0.85, 0.95, 1}
\definecolor{custom_light_pink}{rgb}{1, 0.85, 0.85}
\definecolor{custom_light_green}{rgb}{0.85, 0.98, 0.80}

\title{MultiMed: Multilingual Medical Speech Recognition \\via Attention Encoder Decoder}

\author{Khai Le-Duc$^{1,2,3}$, Phuc Phan$^{4}$, Tan-Hanh Pham$^{5}$, Bach Phan Tat$^{6}$,\\ {\bf Minh-Huong Ngo$^7$, Chris Ngo$^{3}$, Thanh Nguyen-Tang$^{8}$, Truong-Son Hy$^{9}$}\\
$^1$University of Toronto, Canada \hspace{0.2cm}
$^2$University Health Network, Canada \hspace{0.2cm}\\ 
$^3$Knovel Engineering Lab, Singapore \hspace{0.2cm}
$^4$FPT University, Vietnam \hspace{0.2cm}\\
$^5$Florida Institute of Technology, USA \hspace{0.2cm}
$^6$KU Leuven, Belgium \hspace{0.2cm}\\
$^7$VNU University of Engineering and Technology, Vietnam\\
$^8$Johns Hopkins University, USA \hspace{0.2cm}
$^9$University of Alabama at Birmingham, USA\\
\mail \texttt{duckhai.le@mail.utoronto.ca}\\
\github \colorbox{custom_light_green}{\texttt{ \href{https://github.com/leduckhai/MultiMed/tree/master/MultiMed}{https://github.com/leduckhai/MultiMed/tree/master/MultiMed}}}\\}

\begin{document}
\maketitle
\begin{abstract}
Multilingual automatic speech recognition (ASR) in the medical domain serves as a foundational task for various downstream applications such as speech translation, spoken language understanding, and voice-activated assistants. This technology improves patient care by enabling efficient communication across language barriers, alleviating specialized workforce shortages, and facilitating improved diagnosis and treatment, particularly during pandemics. In this work, we introduce \textit{MultiMed}, the first multilingual medical ASR dataset, along with the first collection of small-to-large end-to-end medical ASR models, spanning five languages: Vietnamese, English, German, French, and Mandarin Chinese. To our best knowledge, \textit{MultiMed} stands as the world’s largest medical ASR dataset across all major benchmarks: total duration, number of recording conditions, number of accents, and number of speaking roles. Furthermore, we present the first multilinguality study for medical ASR, which includes reproducible empirical baselines, a monolinguality-multilinguality analysis, Attention Encoder Decoder (AED) vs Hybrid comparative study and a linguistic analysis. We present practical ASR end-to-end training schemes optimized for a fixed number of trainable parameters that are common in industry settings. All code, data, and models are available online.
\end{abstract}


\thispagestyle{plain}
\pagestyle{plain}

\section{Introduction}
Automatic speech recognition (ASR) in the medical domain is a critical foundational task, serving a wide range of downstream tasks and applications, including speech translation \cite{mutal2020ellipsis}, electronic health record \cite{kumah2018electronic}, information extraction \cite{selvaraj2020medication}, speech summarization \cite{VietMed_Sum}. This technology improves patient care by automating clinical documentation \cite{hodgson2016risks}, mitigating shortages of specialized healthcare personnel \cite{latif2020speech}, and contributing to more accurate diagnosis and treatment \cite{luo2024assessing}, particularly under the increased demands observed during pandemic scenarios. Furthermore, the size of the ASR market is projected to reach USD 7.14 billion in 2024, with an anticipated compound annual growth rate (CAGR) of 14.24\% from 2024 to 2030, resulting in a market volume of USD 15.87 billion by 2030 \cite{asr_marketsize}. 

Recent research on ASR in the medical domain has been hindered by the lack of publicly available datasets, mainly due to privacy concerns. Existing datasets (see Table \ref{table:data_literature}), such as the English medical ASR dataset by \citet{fareez2022dataset}, are limited to simulated data on respiratory diseases, restricting research to this category and reducing applicability to diverse accents. The \textit{PriMock57} dataset, containing 57 simulated primary care consultations (9 hours of recordings), also lacks generalizability \cite{korfiatis2022primock57}. The \textit{AfriSpeech-200} dataset \cite{olatunji2023afrispeech} mixes general and medical-domain speech, while the \textit{myMediCon} dataset \cite{htun2024mymedicon} includes Burmese read speech, both of which lack real-world applicability. The \textit{VietMed} dataset \cite{vietmed_dataset} is a real-world dataset focused on the Vietnamese language.

Furthermore, commercial medical ASR APIs, such as Google Cloud Healthcare, IBM Watson, Microsoft Azure Speech Service, Deepgram, and Nuance Dragon Medical One, are not free and do not provide publicly available models for fine-tuning or deployment, nor do they disclose training details.

This work aims to democratize medical ASR, making it freely accessible to everyone. Our key contributions are as follows.
\begin{itemize}
    \item We present the \textit{MultiMed} dataset - the first multilingual medical ASR dataset - which includes human-annotated high-quality real-world medical domain speech in 5 languages. To our best knowledge, \textit{MultiMed} is the world's largest medical ASR dataset on all major diversity benchmarks: total duration (150 hours), number of recording conditions (10), number of accents (16) and number of speaking roles (6). 
    \item We release the first publicly available multilingual medical ASR models, spanning small to large end-to-end configurations. 
    \item We present the first multilinguality study for medical ASR, which includes: reproducible empirical baselines, a monolinguality-multilinguality analysis, Attention Encoder Decoder (AED) vs Hybrid study and a linguistic analysis
    \item We present practical ASR end-to-end training schemes optimized for a fixed number of trainable parameters that are common in industry settings
\end{itemize}

All code, data, and models are published online.

\section{Data}

\subsection{Data Collection}
\label{sec:Data_Collection}
Speech data with human-annotated transcripts were initially collected from real-world medical conversations published by professional medical channels on YouTube. In contrast to simulated datasets in the literature where doctors and patients play roles, our real-world dataset encompasses natural conversations of 10 distinct recording conditions (Documentary, Interview, Lecture, News, Podcast, Webinar, Speech, Talk, Vlog, Workshop) and 6 speaker roles (Lecturer, Doctor, Host, Patient, Podcaster, Broadcaster). Details of data collection for each language to ensure diversity are described in the Appendix \ref{sec:details_data_collection_perLanguage}.

Our adherence to the Fair Use Policy and regulations regarding data consent, privacy, and anonymization of speaker identities in medical research is detailed in the Appendix \ref{sec:ethical_statements}. 

\subsection{Data Quality Control}
Quality control of the initial human-annotated transcripts from professional YouTube channels was carried out through manual review by our annotators, involving the correction of small inaccuracies or the exclusion of too erroneous transcripts. All transcripts were reviewed by medical experts with a certified linguistic level, which ensured, to the best of our knowledge, the final high-quality transcripts. Details of our annotators are described in the Appendix \ref{sec:details_data_quality_control}. Data processing was also performed to further enhance the quality of the transcripts, as described in the Appendix \ref{sec:data_processing}.

\subsection{Data Statistics}
\begin{table*}[t]
\centering
\begin{adjustbox}{width=\textwidth}
\begin{tabular}{l|c|ccccccc}
\hline
\textbf{Dataset} & \textbf{Venue} & \textbf{Dur.} & \textbf{Language} & \textbf{Nature} & \textbf{\#Rec. Cond.} & \textbf{\#Spk} & \textbf{\#Acc} & \textbf{\#Roles} \\ \hline
MultiMed (ours) & - & 150h & Multiling. & Real-world & 10 & 198 & 16 & 6 \\
VietMed \cite{vietmed_dataset} & LREC-COLING & 16h & Vietnamese & Real-world & 8 & 61 & 6 & 6 \\
PriMock57 \cite{korfiatis2022primock57} & ACL & 9h & English & Simulated & 1 & 64 & 4 & 2 \\
Work by \citet{fareez2022dataset} & Nature & 55h & English & Simulated & 1 & N/A & 1 & 2 \\
AfriSpeech-200 \cite{olatunji2023afrispeech} & TACL & $\approx$123h & African English & Read speech & 1 & N/A & N/A & 1 \\
myMediCon \cite{htun2024mymedicon} & LREC-COLING & 11h & Burmese & Read speech & 1 & 12 & 5 & 2 \\ \hline
\end{tabular}
\end{adjustbox}
\caption{Dataset statistics in comparison with all existing works from left to right: Total duration in hours (h), language, nature of speech, number of recording conditions, number of speakers, number of accents, speaking roles. Full details are in Table \ref{table:data_literature} in the Appendix.}
\label{table:data_literature_CutVersion}
\end{table*}

\begin{table}[h]
\centering
\begin{adjustbox}{width=0.475\textwidth}
\begin{tabular}{l|lccc}
\toprule
\textbf{Language}            & \textbf{Set} & \textbf{Samples} & \textbf{Total Dur. (h)} & \textbf{Avg. length (s)} \\ \hline
\multirow{3}{*}{Vietnamese} & Train     & 4548              & 7.81 & 6.19 \\ 
                            & Dev  & 1137              & 1.94 & 6.15 \\ 
                            & Test      & 3437              & 6.02 & 6.31 \\ \midrule
                            
\multirow{3}{*}{English}    & Train     & 27922             & 83.87 & 10.81 \\ 
                            & Dev  & 3082              & 8.96 & 10.46 \\ 
                            & Test      & 5016              & 15.91 & 11.42 \\ \midrule
                            
\multirow{3}{*}{French}     & Train     & 1725              & 5.46 & 11.41 \\ 
                            & Dev  & 52                & 0.18 & 12.13 \\ 
                            & Test      & 358               & 1.15 & 11.57 \\ \midrule
                            
\multirow{3}{*}{Chinese}    & Train     & 1346              & 5.02 & 13.43 \\ 
                            & Dev  & 97                & 0.34 & 12.75 \\ 
                            & Test      & 231               & 0.85 & 13.21 \\ \midrule
                            
\multirow{3}{*}{German}     & Train     & 1551              & 5.37 & 12.46 \\ 
                            & Dev  & 310               & 1.05 & 12.15 \\ 
                            & Test      & 1242              & 4.32 & 12.53 \\ \bottomrule
\end{tabular}
\end{adjustbox}
\caption{Statistics of our data samples: Total duration in hours (h) and average audio length in seconds (s).
}
\label{table:dataset_statistics}
\end{table}

Table \ref{table:data_literature_CutVersion} shows the dataset statistics of our \textit{MultiMed} dataset in comparison with all existing publicly available medical ASR datasets, to the best of our knowledge. As shown in the table, our \textit{MultiMed} dataset is the world's largest medical ASR dataset across all major diversity benchmarks: total duration (150 hours of recordings), number of recording conditions (10), number of accents (16) and number of speaking roles (6).

The statistics for the dataset split for each language are also shown in Table \ref{table:dataset_statistics}.


\section{Problem Definition}
An ASR model transcribes an audio signal into text by mapping an audio signal $x^{T}_{1} := x_{1}, x_{2}, ..., x_{T}$ of length $T$ to the most likely word sequence $w^{N}_{1}$ of length $N$. The relation $w^{*}$ between the acoustic and word sequence is defined as the probability $p$:
\begin{equation}
w^{*} = \operatorname{arg}\max_{w_1^N} \, p(w_{1}^{N}|x_{1}^{T})    
\end{equation}

In beam search process, the auxilary quantity $Q$ for each unknown partial string (tree of partial hypotheses) $w_{1}^{n}$ is described as:
\begin{equation}
\begin{split}
Q(n; w_{1}^{n}) :&= \prod_{n'=1}^{n} p(w_{n'}|w_{0}^{n'-1},x_{1}^{T})\\
&= p(w_{n}|w_{0}^{n-1},x_{1}^{T}) \cdot Q(n-1, w_{1}^{n-1}).
\end{split}
\end{equation}

After eliminating the less likely hypotheses in the beam search process, the word sequence probability is determined by the most optimal hypothesis:
\begin{equation}
p(w_{1}^{N}|x_{1}^{T}) = Q(N; w_{1}^{N}).    
\end{equation}

The complete mathematical formulation of AED is shown in Appendix \ref{sec:AED}, while the formulation for the hybrid model is presented in Appendix \ref{sec:hybrid_wav2vec2}.

\section{Experimental Setups}

\subsection{Model Selection and Training}
\label{sec.model_and_training}

We opted to evaluate the performance of four pre-trained Whisper models \cite{radford2023robust} with varying sizes: Tiny, Base, Small, and Medium. These models, pre-trained on 680,000h of labeled multilingual data, offered a trade-off between accuracy and computational cost, allowing us to explore the impact of model size on performance. Details of hyperparameter tuning are shown in the Appendix \ref{sec:hyperparam_tuning}.

To investigate the impact of different fine-tuning strategies, we explored two main fine-tuning approaches for each model size: \textbf{Decoder-only fine-tuning} (encoder freezing) and \textbf{Fully encoder-decoder fine-tuning}. In the first approach, we focused on fine-tuning only the decoder of the pre-trained Whisper model. The encoder, responsible for aligning audio features, remained frozen during fine-tuning. This strategy aimed to leverage the previously learned representations of the pre-trained encoder for efficient time-frame alignment while adapting learnable parameters in the decoder for vocabulary generation. Otherwise, in the second approach, all parameters in both the encoder and decoder components of the pre-trained Whisper models were learnable. This approach allowed the model to proactively align time-frames in our dataset, potentially leading to better overall performance. The number of parameters for two fine-tuning settings is shown in Table \ref{table:model_configuration}.

\begin{table}[h]
\centering
\begin{adjustbox}{width=0.475\textwidth}
\begin{tabular}{l|cc}
\hline
\textbf{Model} & \textbf{Fully encoder-decoder ft.} & \textbf{Decoder-only ft.} \\ \hline
Tiny     & 37.76M                     & 29.55M                         \\
Base     & 72.59M                     & 52.00M                         \\
Small    & 241.73M                    & 153.58M                        \\
Medium   & 763.86M                    & 456.64M                        \\ \hline
\end{tabular}
\end{adjustbox}
\caption{Statistics of total trainable parameters in the Whisper models for 2 settings: Fully encoder-decoder fine-tuning and decoder-only fine-tuning.}
\label{table:model_configuration}
\end{table}

\subsection{Evaluation Metrics}
To assess the performance of the ASR models, we employed two standard evaluation metrics: Word Error Rate (WER) and Character Error Rate (CER). The description of the two metrics is shown in the Appendix \ref{sec:details_eval_metrics}.

\section{Experimental Results}
\label{sec:experimental_results}
\subsection{Monolingual Fine-tuning}
\begin{table*}[ht]
\centering
\renewcommand{\arraystretch}{1.15}
\begin{adjustbox}{width=\textwidth}
\begin{tabular}{l|cccc|cccc|cccc|cccc}
\hline
\multirow{3}{*}{\textbf{Language}} & \multicolumn{4}{c|}{\textbf{Tiny}} & \multicolumn{4}{c|}{\textbf{Base}} & \multicolumn{4}{c|}{\textbf{Small}} & \multicolumn{4}{c}{\textbf{Medium}} \\ \cline{2-17} 
 & \multicolumn{2}{c}{\textbf{WER}} & \multicolumn{2}{c|}{\textbf{CER}} & \multicolumn{2}{c}{\textbf{WER}} & \multicolumn{2}{c|}{\textbf{CER}} & \multicolumn{2}{c}{\textbf{WER}} & \multicolumn{2}{c|}{\textbf{CER}} & \multicolumn{2}{c}{\textbf{WER}} & \multicolumn{2}{c}{\textbf{CER}} \\ \cline{2-17} 
 & dev & test & dev & test & dev & test & dev & test & dev & test & dev & test & dev & test & dev & test \\ \hline
Vietnamese & 34.23 & 46.98 & 26.88 & 33.04 & 27.16 & 37.74 & 21.20 & 27.34 & 21.82 & 28.77 & 17.97 & 21.81 & 20.05 & 25.43 & 16.77 & 19.87 \\
English & 29.30 & 29.73 & 23.70 & 19.51 & 24.26 & 25.43 & 18.71 & 18.23 & 19.76 & 20.52 & 15.36 & 17.56 & 19.01 & 19.41 & 14.49 & 15.91 \\
French & 54.17 & 52.89 & 34.86 & 34.27 & 43.91 & 42.57 & 27.47 & 27.88 & 35.99 & 33.02 & 24.52 & 22.18 & 34.89 & 31.05 & 24.12 & 21.24 \\
German & 29.38 & 28.22 & 17.29 & 20.00 & 24.27 & 23.09 & 14.65 & 17.16 & 21.68 & 19.91 & 13.58 & 15.96 & 18.90 & 17.92 & 12.07 & 14.57 \\
Chinese & 91.36 & 95.97 & 34.20 & 43.71 & 85.66 & 89.73 & 27.63 & 38.02 & 80.35 & 88.50 & 23.95 & 34.28 & 79.17 & 86.52 & 26.11 & 35.82 \\ \hline
\end{tabular}
\end{adjustbox}
\caption{Main baselines - WERs and CERs of \textbf{decoder-only fine-tuning} (freezing the entire encoder) using different Whisper models on each separate language (\textbf{monolingual fine-tuning})}
\label{tab.metric-wer-cer1}
\end{table*}

\begin{table*}[h]
\centering
\renewcommand{\arraystretch}{1.15}
\begin{adjustbox}{width=\textwidth}
\begin{tabular}{l|cccc|cccc|cccc|cccc}
\hline
\multirow{3}{*}{\textbf{Language}} & \multicolumn{4}{c|}{\textbf{Tiny}} & \multicolumn{4}{c|}{\textbf{Base}} & \multicolumn{4}{c|}{\textbf{Small}} & \multicolumn{4}{c}{\textbf{Medium}} \\ \cline{2-17} 
 & \multicolumn{2}{c}{\textbf{WER}} & \multicolumn{2}{c|}{\textbf{CER}} & \multicolumn{2}{c}{\textbf{WER}} & \multicolumn{2}{c|}{\textbf{CER}} & \multicolumn{2}{c}{\textbf{WER}} & \multicolumn{2}{c|}{\textbf{CER}} & \multicolumn{2}{c}{\textbf{WER}} & \multicolumn{2}{c}{\textbf{CER}} \\ \cline{2-17} 
 & dev & test & dev & test & dev & test & dev & test & dev & test & dev & test & dev & test & dev & test \\ \hline
Vietnamese & 26.79 & 43.32 & 20.18 & 31.06 & 23.69 & 36.48 & 18.73 & 26.18 & 20.61 & 30.27 & 16.94 & 22.55 & 20.73 & 29.81 & 17.25 & 22.59 \\
English & 32.14 & 29.73 & 21.50 & 19.41 & 27.98 & 25.09 & 18.92 & 16.42 & 25.88 & 23.25 & 17.51 & 15.21 & 27.05 & 25.65 & 18.12 & 16.64 \\
French & 55.79 & 55.39 & 34.31 & 35.77 & 45.52 & 44.15 & 27.81 & 28.92 & 43.18 & 42.92 & 30.45 & 29.04 & 44.21 & 41.40 & 29.57 & 28.02 \\
German & 30.81 & 31.29 & 18.72 & 18.43 & 27.93 & 25.25 & 17.15 & 15.11 & 26.16 & 24.64 & 15.74 & 15.46 & 26.22 & 24.13 & 16.02 & 14.68 \\
Chinese & 92.93 & 98.85 & 34.00 & 50.94 & 86.05 & 94.58 & 30.64 & 42.75 & 86.44 & 92.44 & 27.85 & 39.71 & 89.78 & 94.08 & 30.19 & 40.97 \\ \hline
\end{tabular}
\end{adjustbox}
\caption{Main baselines - WERs and CERs of \textbf{fully encoder-decoder fine-tuning} using different Whisper models on each separate language (\textbf{monolingual fine-tuning})}
\label{tab.metric-wer-cer2}
\end{table*}

We fine-tuned various variants of the Whisper model in each language separately (known as monolingual fine-tuning) and analyzed the impact of model size and transfer learning  (decoder-only vs. full encoder-decoder fine-tuning) on recognition accuracy, as shown in Table \ref{tab.metric-wer-cer1} and \ref{tab.metric-wer-cer2}.

A clear correlation was observed between the size and performance of the model. As the model size increased from Tiny to Medium, WER and CER generally decreased across all languages, indicating that larger models better capture complex audio-text representations, improving accuracy.

The best results for most languages were obtained by fine-tuning only the decoder of the Medium model: Vietnamese achieved 20.05\% and 25.43\% WERs on the dev and test sets, respectively; English reached 19.01\% and 19.41\% WERs; and French yielded 34.89\% and 31.05\% WERs.

An exception was Chinese, where fine-tuning the Small model's decoder produced the best results: 23.95\% and 34.28\% CERs on the dev and test sets. Since Chinese uses characters as fundamental units of meaning, CER is a more accurate measure of recognition than WER \cite{wang2016oc16, gao2006just}, unlike alphabetic languages.

\subsection{Multilingual Fine-tuning}
In addition to fine-tuning each language separately, we also combined all languages for experimentation, known as multilingual fine-tuning, as shown in Table \ref{tab.result-multidata}. In multilingual fine-tuning, we achieved superior performance in most languages, though there was a slight performance degradation for Chinese, compared to monolingual fine-tuning in Table \ref{tab.metric-wer-cer2}. Both high-resource languages, such as English, and lower-resource languages, including Vietnamese, French, and German, showed improvement under the multilingual fine-tuning regime. This outcome is noteworthy, as previous studies on multilingual fine-tuning observed that shared discrete latent speech representations across languages such as Vietnamese, English, Chinese, French, and German tend to cluster at large distances, and therefore usually affect accuracy in the multilingual setting \cite{facebook2020wav2vec2, conneau21_XLSR53, vieting2023efficient, tuske2014data, chuangsuwanich2016multilingual}.

\begin{table}[h]
\centering
\begin{adjustbox}{width=0.425\textwidth}
\begin{tabular}{l|cccc}
\hline
\multirow{2}{*}{\textbf{Language}} & \multicolumn{2}{c}{\textbf{WER}} & \multicolumn{2}{c}{\textbf{CER}} \\ \cline{2-5} 
 & dev & test & dev & test \\ \hline
Vietnamese & 23.11 & 30.22 & 18.78 & 22.51 \\
English & 18.92 & 16.62 & 12.97 & 11.05 \\
French & 43.62 & 37.27 & 29.24 & 24.25 \\
German & 25.26 & 22.92 & 15.31 & 14.05 \\
Chinese & 89.78 & 101.97 & 26.65 & 41.21 \\ \hline
\end{tabular}
\end{adjustbox}
\caption{Main baselines - WERs and CERs of \textbf{fully encoder-decoder fine-tuning} using Small Whisper model on all languages (\textbf{multilingual fine-tuning})}
\label{tab.result-multidata}
\end{table}

\subsection{AED vs Hybrid}
\begin{table*}[h]
\centering
\resizebox{0.6\textwidth}{!}{%
\begin{tabular}{ll|cccc}
\hline
\multicolumn{2}{l|}{\multirow{2}{*}{}} & \multicolumn{2}{c}{\textbf{AED}} & \multicolumn{2}{c}{\textbf{Hybrid}} \\ \cline{3-6} 
\multicolumn{2}{l|}{} & \textbf{Small} & \textbf{Medium} & \textbf{w2v2-Viet} & \textbf{XLSR-53-Viet} \\ \hline
\multicolumn{1}{l|}{\multirow{2}{*}{\textbf{WER}}} & \textbf{dev} & 21.8 & 20.1 & 25.9 & 25.7 \\
\multicolumn{1}{l|}{} & \textbf{test} & 28.8 & 25.4 & 29.0 & 28.8 \\ \hline
\multicolumn{2}{l|}{\textbf{\#Data}} & \multicolumn{2}{c}{\begin{tabular}[c]{@{}c@{}}680,000h\\ labeled multiling.\\ (691h labeled Viet.)\end{tabular}} & \begin{tabular}[c]{@{}c@{}}1200h\\ unlabeled Viet.\end{tabular} & \begin{tabular}[c]{@{}c@{}}56,000h\\ unlabeled multiling.\\ +1200h\\ unlabeled Viet.\end{tabular} \\ \hline
\multicolumn{2}{l|}{\textbf{\#Params}} & 153M & 456M & 123M & 123M \\ \hline
\multicolumn{2}{l|}{\textbf{\#Layers}} & 12 & 24 & 8 & 8 \\ \hline
\multicolumn{2}{l|}{\textbf{Width}} & 768 & 1024 & 768 & 768 \\ \hline
\multicolumn{2}{l|}{\textbf{\#Att. Heads}} & 12 & 16 & 16 & 16 \\ \hline
\multicolumn{2}{l|}{\textbf{Features}} & \multicolumn{2}{c}{MFCC} & \multicolumn{2}{c}{Raw waveform} \\ \hline
\multicolumn{2}{l|}{\textbf{LM fusion}} & \multicolumn{2}{c}{Deep fusion} & \multicolumn{2}{c}{Shallow fusion} \\ \hline
\end{tabular}%
}
\caption{Comparison between AED and Hybrid experiments. WERs are reported on our Vietnamese dev and test set. All models were fine-tuned on the same Vietnamese set. Hybrid models employ wav2vec 2.0 as acoustic model \cite{facebook2020wav2vec2}. Full details of experiments are shown in Appendix \ref{sec.extra_experiments_AED_Hybrid} and the breakdown per speaker is shown in Table \ref{tab.breakdown_HybridASR_perSpeaker} in the Appendix.}
\label{tab.AED_vs_Hybrid}
\end{table*}
End-to-end ASR, with the AED approach, and Hybrid ASR models (Hidden Markov Models) are two key paradigms in ASR research. This section compares AED and Hybrid ASR models. For a fair comparison, we use wav2vec 2.0 \cite{facebook2020wav2vec2} as the acoustic model for Hybrid ASR, as it is a Transformer-based encoder, similar to the Transformer-based encoder-decoder of Whisper.

Table \ref{tab.AED_vs_Hybrid} presents a comparison between the AED and Hybrid models. The AED models were pre-trained on 680,000 hours of labeled multilingual data, including 691 hours of Vietnamese, while the Hybrid models were pre-trained on unlabeled data. Despite having fewer parameters and less labeled data, Hybrid models achieve comparable WERs on the Vietnamese test set. AED models only outperform Hybrid models significantly when scaled three times. This finding supports prior research on the data and computational efficiency of Hybrid models in general-domain ASR \cite{luescher2019:librispeech, zeyer2018:asr-attention, zeyer2018:attanalysis, zeyer2019comparison}, and is the first confirmation of this trend in the medical domain.


\section{Ablation Study: Freezing Schemes}
\begin{table*}[h]
\resizebox{\textwidth}{!}{%
\begin{tabular}{l|cccc|cccc|cccc|}
\hline
\multirow{3}{*}{\textbf{Language}} & \multicolumn{4}{c|}{\textbf{0-8 encoder}} & \multicolumn{4}{c|}{\textbf{3-11 encoder}} & \multicolumn{4}{c|}{\textbf{0-8 encoder \& 0-8 decoder}} \\ \cline{2-13} 
 & \multicolumn{2}{c}{\textbf{WER}} & \multicolumn{2}{c|}{\textbf{CER}} & \multicolumn{2}{c}{\textbf{WER}} & \multicolumn{2}{c|}{\textbf{CER}} & \multicolumn{2}{c}{\textbf{WER}} & \multicolumn{2}{c|}{\textbf{CER}} \\ \cline{2-13} 
 & dev & test & dev & test & dev & test & dev & test & dev & test & dev & test \\ \hline
Vietnamese & 21.27 & 29.32 & 17.60 & 22.07 & 21.28 & 30.74 & 17.60 & 22.97 & 23.44 & 33.30 & 19.33 & 24.78 \\
English & 25.68 & 26.50 & 14.87 & 17.84 & 22.68 & 25.20 & 14.73 & 16.90 & 16.78 & 32.11 & 12.78 & 22.42 \\
French & 39.36 & 35.50 & 27.48 & 23.70 & 38.71 & 35.03 & 26.32 & 23.59 & 37.68 & 35.93 & 25.69 & 24.02 \\
German & 23.65 & 21.49 & 15.04 & 13.64 & 22.82 & 20.94 & 14.30 & 13.29 & 22.64 & 23.04 & 14.54 & 15.14 \\
Chinese & 78.97 & 88.33 & 23.37 & 35.72 & 83.49 & 89.48 & 25.20 & 37.07 & 80.75 & 94.91 & 28.32 & 38.80 \\ \hline
 & \multicolumn{4}{c|}{\textbf{0-8 encoder \& 3-11 decoder}} & \multicolumn{4}{c|}{\textbf{0-11 encoder \& 0-8 decoder}} & \multicolumn{4}{c|}{\textbf{3-11 encoder \& 3-11 decoder}} \\ \hline
Vietnamese & 34.98 & 32.81 & 29.34 & 24.65 & 24.75 & 32.11 & 20.86 & 25.06 & 40.87 & 32.10 & 36.06 & 24.30 \\
English & 20.61 & 28.31 & 15.55 & 19.56 & 16.06 & 31.32 & 12.68 & 22.34 & 21.53 & 34.81 & 17.09 & 22.96 \\
French & 35.04 & 40.70 & 23.32 & 32.96 & 37.97 & 37.39 & 27.25 & 26.60 & 57.26 & 40.10 & 44.82 & 28.83 \\
German & 22.22 & 21.02 & 13.83 & 13.35 & 22.11 & 22.26 & 14.65 & 14.98 & 22.86 & 22.47 & 15.01 & 15.23 \\
Chinese & 79.76 & 93.51 & 23.93 & 35.34 & 84.67 & 87.84 & 26.24 & 34.36 & 132.80 & 103.04 & 53.74 & 41.21 \\ \hline
\end{tabular}%
}
\caption{Ablation study - WERs and CERs of various freezing schemes using Small Whisper model on each separate language (\textbf{monolingual fine-tuning}). Small Whisper model has 12 layers in the encoder and 12 layers in the decoder. For example, \textit{0-8 encoder} means freezing all layers from layer 0 to layer 8 in the encoder, the rest layers are fine-tuned.
}
\label{tab.ablation_freezing_fullTable}
\end{table*}
This section presents the results of our ablation study. The Small Whisper models fit within a 24GB GPU without out-of-memory issues. We evaluated the impact of freezing various layers on performance, focusing on test set WERs for most languages and CERs for Chinese. The tested freezing configurations are shown in Table \ref{tab.ablation_freezing_fullTable}.

In both the \textit{0-8 encoder} and \textit{3-11 encoder} settings, model performance on test sets is worse than when the entire encoder is frozen in Table \ref{tab.metric-wer-cer2}. This suggests that, within a fixed budget, freezing the entire encoder, which aligns time frame features with language representations, is crucial to achieve high accuracy and computational efficiency, as seen in the general domain AED ASR \cite{ueno2018encoder}.

We also explored the effect of freezing Whisper's decoder, focusing on fine-tuning only the last three layers of both the encoder and decoder (\textit{0-8 encoder \& 0-8 decoder}). As shown in Table \ref{tab.ablation_freezing_fullTable}, this setup resulted in worse performance compared to fine-tuning only the decoder while freezing layers 0-8 of the encoder (\textit{0-8 encoder}). Performance degradation likely results from a significant reduction in trainable parameters in the decoder, which is responsible for generating subword units. Given the fixed vocabulary, out-of-vocabulary (OOV) words, and context length in Whisper's Byte-Pair Encoding (BPE) tokenizer \cite{gage1994new}, the decrease in trainable autoregressive parameters likely hinders the decoder's ability to effectively separate subword tokens, leading to reduced decoding accuracy \cite{ho2024block, bapna2020controlling}.

We fine-tuned the first three decoder layers and the last three encoder layers (\textit{0-8 encoder \& 3-11 decoder}), which generally resulted in higher test set accuracy for most languages compared to \textit{0-8 encoder \& 0-8 decoder}. This suggests that freezing a contiguous set of layers is the key to achieving high accuracy with an equivalent number of trainable parameters in the decoder.

Fine-tuning the last three decoder layers (\textit{0-11 encoder \& 0-8 decoder}) also outperformed \textit{0-8 encoder \& 0-8 decoder} in test accuracy and was competitive with \textit{0-8 encoder \& 3-11 decoder}, despite fewer trainable parameters. Likewise, the \textit{3-11 encoder \& 3-11 decoder} configuration yielded the worst performance in all languages. These findings support the hypothesis that consistent freezing of contiguous layer groups is critical for high accuracy within a fixed parameter budget.

\section{Error Analysis}
To our best knowledge, there has been no error analysis based on the linguistic perspective for languages other than English. Therefore, we used the English literature to compare with our findings.

We manually analyzed the errors in 50 randomly collected samples from each language. Generally, the errors observed in medical ASR systems are diverse and cover a wide range of issues. For all 5 languages, these typically include misrecognition of drug names and dosages, incorrect medical institutions, anatomical discrepancies (e.g., left-right confusion), medical terms' inconsistencies, mismatches in patient age and gender, incorrect identification of physician names, and inaccuracies in dates. These findings are consistent with the study by \citet{hodgson2016risks} in the English medical ASR dataset. Additionally, the misrecognition is exacerbated by the generation of non-existent terms, which is also known as hallucination in the Large Language Models (LLMs) era, as well as omissions (e.g. deletion errors) and duplications (e.g., insertion errors) within the ASR output (see Figure \ref{ASR_errors_example} in the Appendix). These findings are also confirmed by \citet{mcgurk2008effect} in English ASR for radiology reports.

Furthermore, ASR errors typically arise from the proximity of vowels in the phonological space for Vietnamese, English, German, and French, while for Chinese, confusion predominantly stems from minimal pairs with distinct tones and homophones. Detailed error analysis based on the linguistic perspective for each language is in Appendix \ref{sec:error_analysis_linguistic_perspective}.

\section{Conclusion}
In this work, we present \textit{MultiMed}, a real-world dataset for ASR in the medical domain, accompanied by a collection of small-to-large end-to-end ASR models, covering five languages: Vietnamese, English, German, French, and Mandarin Chinese. To our best knowledge, \textit{MultiMed} stands as the world's largest medical ASR dataset across all major benchmarks. 

As the first study of multilingual ASR in the medical domain, our findings demonstrate that \underline{\textbf{(1):}} multilingual fine-tuning produces superior accuracy compared to monolingual fine-tuning, although shared discrete latent speech representations across languages, such as Vietnamese, English, Chinese, French and German, exhibit clustering at large distances, which could potentially reduce accuracy in a multilingual fine-tuning setting. 
Furthermore, in the AED vs Hybrid study, we showed that \underline{\textbf{(2):}} Hybrid models remain more efficient in terms of data utilization and computational performance compared to AED models. In the layer-wise ablation study of AED models, we found that \underline{\textbf{(3):}} on a fixed budget, freezing the entire encoder  is important for achieving both high accuracy and computational efficiency. Additionally, \underline{\textbf{(4):}} maintaining the consistent freezing of a contiguous group of layers is important for achieving high accuracy. Finally, as shown in the linguistic analysis for multilingual medical ASR, we observed that \underline{\textbf{(5):}} medical ASR errors often involve misrecognitions of drug names, dosages, institutions, anatomical details, demographics of patients, physician names, etc., along with hallucinated terms, omissions, and duplications. \underline{\textbf{(6):}} Errors also often arise from the proximity of vowels in the phonological space for Vietnamese, English, German and French, while for Chinese, confusion predominantly stems from minimal pairs with distinct tones and homophones.

\section{Limitations}
\textbf{Open research questions}: Several research questions about the impact of multilinguality on medical ASR remain unaddressed and fall outside the scope of this study.
\begin{itemize}
    \item Cross-language transfer learning: How can transfer learning be optimized to leverage data from high-resource languages to improve medical ASR performance in low-resource languages? Can shared acoustic and linguistic representations (e.g., from hospitals' recording conditions and shared medical terms across languages) effectively bridge the gap between typologically different languages?
    \item Zero-shot and few-shot medical ASR: What are the best methods for enabling general-domain ASR models to understand unseen medical-domain test set (zero-shot learning) or to adapt with minimal medical-domain data (few-shot learning)? How can medical-domain models be trained to generalize effectively across languages without overfitting to dominant languages (e.g., English) in the dataset?
    \item Code-Switching Challenges: How does each ASR module handle code-switching, where speakers switch between two or more languages within the same sentence, especially for medical terms?
    \item Bias and Fairness in Multilingual Medical ASR: How can we address biases in multilingual medical ASR models that disproportionately affect minority languages or speakers with diverse accents, especially when patients and doctors are not of major ethnicity? What metrics and evaluation protocols should be established to assess fairness and inclusivity in multilingual medical ASR systems?
\end{itemize}

\noindent\textbf{Clinical impact}: The primary objective of our study is to establish baselines rather than introduce novel techniques to minimize WER in medical ASR systems. Given the critical nature of medical transcription, inaccuracies in ASR output can have serious implications, potentially affecting patient diagnoses and treatment decisions \cite{adane2019role}. Thus, real-world deployment of our systems should be preceded by pilot testing in clinical environments to ensure reliability prior to full-scale implementation.

\bibliography{custom}

\clearpage 


\appendix

\onecolumn
\tableofcontents
\newpage

\twocolumn

\section{Related Works}
\label{related_works}

Among the limited existing studies below, to the best of our knowledge, none of them have made their datasets or pre-trained models publicly available, nor have they been conducted on publicly accessible datasets due to privacy concerns, which poses a significant challenge for the reproducibility and deployment of medical ASR research\footnote{Medical ASR, also known as Medical-domain ASR, focuses on developing ASR systems specifically tailored for healthcare environments, such as hospitals, clinics, and telemedicine. It aims to transcribe medical dictations, conversations between healthcare providers and patients, or interactions with electronic health records (EHRs). The term medical ASR does not refer to the "ASR of pathological speech", which focuses on developing models capable of recognizing and transcribing speech from individuals with speech impairments or disorders. These impairments can be due to conditions such as dysarthria, aphasia, stuttering, or neurological diseases such as Parkinson}.

\textbf{Multilingual medical ASR}: The research carried out by \citet{luescher2022:hykist} has focused on the development of Hybrid ASR systems \cite{luscher19_interspeech} in the RETURNN framework \cite{zeyer2018returnn, doetsch2017returnn} to transcribing multilingual telephone conversations between patients and physicians, using Gammatone features \cite{gammatone_ralf} as input in a supervised only approach. \citet{vieting2023efficient} examine the efficient utilization of the large multilingual acoustic pre-trained model XLSR-53 \cite{conneau21_interspeech} for medical ASR in three languages, focusing on resolving the issue of sampling rate mismatch using wav2vec 2.0 \cite{baevski2020wav2vec} as encoder and RASR \cite{rybach2011rasr} as decoding framework. In \cite{sakti2014towards}, a multilingual acoustic model fine-tuned on in-house medical domain data is presented utilizing weighted finite-state transducers \cite{mohri2002weighted}, speaker adaptive training \cite{anastasakos1996compact}, and  boosted maximum mutual information \cite{povey2008boosted} in conjunction with Kaldi decoding \cite{povey2011kaldi} of n-gram \cite{ney1994structuring} language models for every specific language. However, to the best of our knowledge, in all studies separate monolingual models are typically used for each respective language, rather than utilizing a unified multilingual model capable of transcribing multilingual conversations seamlessly. Therefore, we are the first study to present a unified multilingual model that can dynamically adapt to different languages in medical conversations without the need for separate models.


\textbf{Acoustic challenges for medical ASR}: In this context, several challenges arise, including variability in acoustic and recording conditions, the mismatch in telephony bandwidth, the impact of medical mask usage, and the presence of background noise from various devices and dynamic environmental factors \cite{luescher2022:hykist}. In addition, a bidirectional input issue is observed, as a single recording channel is shared between the physician and the patient in emergency room and hospital settings. Studies such as \cite{edwards2017medicalspeech, chiu2018medconv, kar2021operation, dua2023noise} have addressed the challenges related to difficult acoustic conditions by modifying model components like feature extractor, acoustic model, and so on. Furthermore, studies like \cite{salimbajevs2022automatic} address the robustness in noisy acoustic environments using a large amount of unlabeled medical ASR data. In addition, \citet{luo2024assessing} uses emergency medical services or prehospital care as a research context to generate data in the domain, as it represents a prototypical example of dynamic and variable medical environments, involving numerous participants, such as healthcare professionals, patients, bystanders, and family members.

\textbf{Language modeling for medical ASR}: The specialized medical terminology in each language presents an additional challenge. A simple method to address these challenges involves correcting ASR errors at the output level \cite{mani2020towardsmedical, hsu2024enhanced, mani2020asr} or focusing on medically named entities \cite{afonja2024performant, leduc2024medicalspokennamedentity, suominen2015benchmarking}. Another approach is to train a domain-specific language model to decode the ASR encoder \cite{jiang2021sequence}. Furthermore, LLMs, such as the GPT series \cite{openai2024gpt4technicalreport, brown2020languagemodelsfewshotlearners}, Gemini \cite{team2024gemini, team2023gemini}, and Llama \cite{touvron2023llama2openfoundation, dubey2024llama3herdmodels} for example, have potential utility in rectifying medical ASR errors \cite{adedeji2024sound}. Another study by \citet{sunkara2020robust} involves the joint modeling of punctuation and truecasing in medical ASR transcripts utilizing pre-trained language models, such as BERT \cite{devlin-etal-2019-bert}. 

\textbf{Application of medical ASR}: One of the most common use cases is for clinical documentation \cite{latif2020speech}. It is a laborious and complex task that could lead to burnout of the clinician \cite{arndt2017tethered}, inefficiency of the doctor-patient time \cite{sinsky2016allocation}, and lower patient satisfaction \cite{pelland2017like}. The adoption of electronic health records (EHRs) has been progressively implemented to optimize this process, leveraging medical ASR technology \cite{van2021digital, zhang2023intelligent, johnson2014systematic, saxena2018provider}. Secondly, in the context of emergency medical services, medical ASR has been evaluated for its influence on stroke detection, demonstrating potential to enhance response times and diagnostic accuracy \cite{donnelly2022use}. Thirdly, ASR can be employed in surgical environments to enhance communication between the surgeon and both human assistants (e.g., surgical nurses) and digital systems (e.g., robotic arms) \cite{ruby2020automatic, schulte2020automatic}. Fourthly, in pediatric healthcare, medical ASR systems have been investigated for their potential application in remote care management \cite{nayar2017towards}. Fifthly, medical ASR can be employed to support individuals with hearing impairments or disorders related to voice, speech, or language, facilitating more effective communication \cite{wendt2011assistive}.

\twocolumn
\section{Ethical Statements}
\label{sec:ethical_statements}
Speech data accompanied by high-quality human-annotated transcripts was obtained from YouTube in compliance with the Fair Use Policy and Vietnamese regulations governing data consent, privacy, and medical research, as detailed in this section. 

According to Vietnamese law, which is applicable to the location of the hosting of the data and the site of all research activities, international and local researchers are authorized to collect and use the data exclusively for scientific purposes. To further ensure data privacy, segments of the dataset with the potential to reveal speaker identities were anonymized.

\subsection{Fair Use}
The research adhered rigorously to the principles of Fair Use as defined by the U.S. Copyright Office\footnote{https://www.copyright.gov/fair-use/}, which are also applicable to content on the YouTube platform. Fair Use is governed by Section 107 of the Copyright Act, which provides a legal framework for evaluating whether a specific use of copyrighted material qualifies under this doctrine. The statute identifies several examples of permissible uses, including criticism, commentary, news reporting, teaching, scholarship, and research, which are particularly relevant in the context of academic and scientific work.

Section 107 outlines a multifactorial approach to determining fair use, which requires an assessment of four key factors. These include:
\begin{itemize}
    \item (1) \textbf{Purpose and character of the use, including whether the use is of a commercial nature or is for nonprofit educational purposes:}
    This factor evaluates how the copyrighted work is being utilized, particularly whether the use serves a commercial purpose or is directed toward nonprofit educational objectives. Courts tend to favor claims of fair use when the purpose is educational and non-profit rather than commercial. Furthermore, the concept of "transformative use" plays a significant role in this determination. Transformative uses are characterized by their ability to add new meaning, insight, or purposes to the original work, altering its character in a way that differentiates it from the initial intention. Transformative uses that do not replace or compete with the original purpose of the work are more likely to qualify as fair use.
    \item (2) \textbf{Nature of the copyrighted work:} This factor examines the type of work involved and its relationship to the copyright's goal of fostering creative expression. Works that are highly imaginative or creative, such as novels, films, or songs, receive stronger copyright protection, making their use less likely to be considered fair. In contrast, factual or informational works, such as technical articles or news reports, are less stringently protected, and their use may more readily align with fair-use principles. Additionally, unpublished works are generally given greater protection and their unauthorized use is less likely to meet fair use criteria.
    \item (3) \textbf{Amount and substantiality of the portion used in relation to the copyrighted work as a whole:} 
    This factor assesses both the quantitative and qualitative aspects of the material used in relation to the entire copyrighted work. The use of larger portions of a work typically weighs against fair use, though exceptions exist in cases where the entirety of the work is used for a justified purpose. Conversely, even the use of a small excerpt may be deemed unfair if it constitutes the "heart" or most significant and recognizable aspect of the original work. In this evaluation, the balance between the necessity of the portion used and its impact on the original work is critical.
    \item (4) \textbf{Effect of the use on the potential market for or value of the copyrighted work:} 
    This factor considers the economic impact of the use without a license on both the existing market and the potential future markets for the copyrighted work. Courts analyze whether the unauthorized use undermines the market value or competes with the copyright holder's ability to monetize their work. If the unlicensed use causes substantial harm to the market or diminishes the value of the original work, it is less likely to qualify as fair use.
\end{itemize}

These factors collectively inform the determination of whether the usage is lawful under the doctrine of Fair Use, providing a nuanced and case-specific analysis.

In accordance with applicable legal frameworks, our work is justified under the provisions of the Fair Use doctrine. This assertion is supported by a detailed interpretation of the Fair Use principles \footnote{https://copyrightalliance.org/faqs/what-is-fair-use/} by copyrightalliance.org and the ELRC Report on legal issues in web crawling \footnote{http://www.elra.info/media/filer\_public/2021/02/12/elrc-legal-analysis-webcrawling\_report-v11.pdf} by Pawel Kamocki, which emphasize the transformative nature of our research, its non-commercial scientific purpose and its minimal impact on the market value of the original content. These considerations collectively align with the statutory factors outlined in copyright law, underscoring the legitimacy of our approach. Our detailed interpretation of the Fair Use principles is as follows:

\begin{itemize}
    \item (1) \textbf{Purpose and Character of Use}: The data were collected and utilized strictly for non-commercial and research purposes, aligning with the principles of Fair Use. Rather than directly using the videos obtained from YouTube, we transformed them into audio files at a predefined sampling rate. Long audio files, typically around an hour in duration, were segmented into shorter clips of 10 to 30 seconds. The segments were then randomly shuffled to ensure that they could not be reconstructed to form the original videos. This transformation and randomization process render the dataset distinctly different from the original content, thus qualifying as transformative use. Furthermore, this approach does not substitute for the original purpose or value of YouTube videos.
    \item (2) \textbf{Nature of the Copyrighted Work}: The extracted content primarily consists of factual, non-fictional medical conversations, which further supports its qualification as Fair Use. In addition, YouTube videos are publicly accessible throughout the world, fulfilling the criterion related to the publication status of the copyrighted material.
    \item (3) \textbf{Amount and Substantiality of the Portion Used}: Although there is no quantitative metric to precisely assess the fairness of a specific use, the randomly shuffled 10- to 30-second audio segments do not provide the full context or meaning of the original videos. These short segments are incapable of reproducing or capturing the core or ``heart'' of the copyrighted works.
    \item (4) \textbf{Effect on the Potential Market}: Our dataset does not serve as a competitor to the original content on YouTube. The 10- to 30-second audio segments do not detract from the YouTube viewership or impact the commercial interests of copyright owners. As a result, our work does not interfere with the potential market value of the original videos or undermine the business of copyright owners.
\end{itemize}

By adhering to these principles, we ensure compliance with Fair Use guidelines while maintaining the scientific and ethical integrity of our research. Numerous related works have been conducted that utilize the extraction of video content from YouTube for academic and noncommercial purposes. These studies typically involve systematic retrieval of publicly available videos, followed by their conversion to audio formats to facilitate various lines of research, such as ASR, NLP, and multimedia analysis. Such approaches often aim to leverage the diverse linguistic, cultural, and acoustic features inherent in the vast repository of YouTube content while adhering to ethical guidelines and copyright regulations to ensure the integrity and legality of the research, such as GigaSpeech\footnote{https://github.com/SpeechColab/GigaSpeech} (China \& USA), VoxCeleb\footnote{https://www.robots.ox.ac.uk/~vgg/data/voxceleb/} (UK), VoxLingua107\footnote{https://bark.phon.ioc.ee/voxlingua107/} (UK). 

\subsection{Data Consent}
Our waiver of data consent for the collection of medical ASR datasets is justified based on ethical and regulatory considerations, particularly when the data are deidentified and pose minimal risk to individuals. In compliance with institutional review board (IRB) guidelines and regulatory frameworks, such as the Common Rule, consent can be waived if it is impractical and research has significant potential to advance medical knowledge or improve healthcare outcomes. Anonymization techniques, including speaker de-identification, ensure that patient confidentiality is maintained, mitigating privacy concerns. Additionally, the dataset is used strictly for research purposes, with safeguards in place to prevent misuse or unauthorized access. These measures collectively support the ethical and legal justification for waiving individual data consent while upholding privacy protections.

The publication of research data in this study adheres to relevant legal frameworks concerning data consent and privacy protection, both within Vietnam and internationally. A comprehensive explanation is provided below.
\begin{itemize}
    \item \textbf{Global Data Protection and Privacy Compliance}: Of the 194 countries globally, 137 have adopted Data Protection and Privacy Legislation\footnote{https://unctad.org/page/data-protection-and-privacy-legislation-worldwide}, as documented by the United Nations (UN). This includes key signatories such as the USA, EU member states (e.g., Germany), and Vietnam. In alignment with these international frameworks, Vietnam's Personal Data Protection Act stipulates in Article 6 that "The protection of personal data is carried out in accordance with international treaties to which the Socialist Republic of Vietnam is a member." This establishes that Vietnamese data protection laws comply with international standards, ensuring compatibility and lawful handling of personal data for global collaboration in research.
    \item \textbf{Exemption for Sensitive Data Processing for Research}: Article 20, Section 4 of Vietnam's Personal Data Protection Act explicitly states that "The party processing personal data is not required to register for processing sensitive personal data in the case of research purposes." This provision legally allows researchers to process sensitive data, including medical and speech-related datasets, without the explicit consent of individuals, provided the purpose is confined to scientific inquiry.
    \item \textbf{No Consent Requirement for Data Publication in Research}: Under Article 16 of Vietnam's Personal Data Protection Act, the principle of data deletion is waived for cases involving scientific research, statistics, or legal obligations. The law specifies that: "Data deletion will not apply at the request of the data subject in the following cases: Personal data is processed to serve legal requirements, scientific research, and statistics." Thus, researchers are exempt from obtaining consent from data subjects for the inclusion of their data in publications, reaffirming the permissibility of this study's data handling practices.
    \item \textbf{Encouragement of Research Publication in Vietnam}: The Law on Medical Examination and Treatment, in conjunction with the Constitution of the Socialist Republic of Vietnam, underscores the importance of scientific dissemination. Article 22 mandates that medical practitioners and researchers "are responsible for updating relevant medical knowledge (...) including (...) c) Publish scientific research (...)." This legal encouragement promotes the proactive sharing of findings, particularly when involving sensitive medical data, as part of advancing public health and scientific understanding.
    \item \textbf{Legal Protections for Researchers}: Article 42 of the Law on Medical Examination and Treatment provides explicit protections for researchers. It states that researchers are "protected by the law and not responsible when a medical incident still occurs after complying with regulations." This ensures that any unforeseen outcomes related to the use or publication of research data, provided it aligns with statutory requirements, do not hold researchers liable.
    \item \textbf{Data Collection and Jurisdictional Compliance}: The dataset utilized in this study was collected using Vietnamese IP addresses and a web crawler authorized by a Vietnamese government-recognized company. This method adheres to Vietnam's Cybersecurity Law, as outlined in Article 26 of the Constitution of the Socialist Republic of Vietnam. It mandates that "Domestic and foreign enterprises providing services on telecommunications networks, the Internet, and value-added services in cyberspace in Vietnam have activities of collecting, exploiting, analyzing, and processing information data (...) created by service users in Vietnam must store this data in Vietnam (...) as prescribed by the Government." Consequently, YouTube, as a service provider, must comply with Vietnamese regulations concerning data generated within the country's cyberspace.
    \item \textbf{International Researchers and Cross-Border Legal Alignment}: Articles 2 and 10 of the Vietnamese Civil Code on Civil Relations with Foreign Elements assert the application of Vietnamese law to international civil relations involving foreign researchers. Specifically, the Code emphasizes that "The provisions of Vietnamese civil law apply to civil relations involving foreign elements (...). In case the application or consequences of the application of foreign law are contrary to (...) the Vietnam Civil Code and other basic principles of Vietnamese law, then Vietnamese law applies." This ensures that international researchers working with Vietnamese data must adhere to Vietnamese laws, while simultaneously receiving legal protections and encouragement under these frameworks.
\end{itemize}

The dataset utilized in this study comprises YouTube content predominantly centered on medical themes, including televised medical shows, interviews, and educational lectures. In all cases, the participants in the videos spoke directly to the camera, demonstrating awareness that their content was intended for public dissemination. This awareness comes from the context of these videos, which were explicitly produced with the goal of providing accurate and accessible medical knowledge to YouTube audiences. Importantly, these videos were officially published by reputable national television channels, ensuring a professional standard of production and adherence to broadcasting regulations.

In contrast, YouTube videos created by amateur content creators, where the individuals featured may not have been aware of being recorded or of the eventual publication of the footage, were explicitly excluded from our dataset. This exclusion criterion was implemented to maintain ethical standards, particularly regarding informed consent and privacy. By limiting the dataset to professionally produced content with a clear intention of public dissemination, we aimed to ensure that the data collected adhered to legal and ethical guidelines on participant awareness and data use.

\twocolumn
\section{Details of Data Creation}
\subsection{Details of Data Collection per Language}
\label{sec:details_data_collection_perLanguage}
\subsubsection{English}
The data was collected from YouTube using the 2024 ICD-10-CM Codes to ensure diversity. We searched for diseases associated with the first 8 codes: A00-B99 (Infectious diseases), C00-D49 (Neoplasms), D50-D89 (Blood Diseases), etc. Due to time constraints, we searched only for these codes, applying filters for videos longer than 20 minutes and with subtitles to ensure accuracy. The videos were manually selected, prioritizing diverse speakers, accents, and contexts.

For the first 3 codes, we obtained 20 hours of video and subtitles and 10 hours for the rest. Videos and metadata, including recording conditions, speaker roles, genders, and accents, were saved.

\subsubsection{French}
We collected French medical videos from YouTube using terms such as ``urgence'', ``consultation médicale'', and ``cancer''. The videos required Closed Captions (CC) with timing, either manually annotated or auto-generated by YouTube. We focused on videos over 20 minutes, covering topics such as oncology, cardiology, and pediatrics, from diverse contexts (lectures, interviews, consultations) and recording conditions (clean audio to noisy emergency rooms with multiple speakers) and prosodies (calm narration to distressed cries).

\subsubsection{Chinese}
We tried collecting Chinese videos using the same method as for French, but found very few Mandarin videos with CC from mainland China. Most Chinese subtitles were hardcoded, and available CC were in English (from Singapore) or Traditional Chinese (from Taiwan or Hong Kong). After attempting to upload Chinese videos to our channel for automatic CC generation, we found YouTube could not generate subtitles due to language complexity. As a result, we mainly used videos from Singapore and Taiwan, with fewer from mainland China.

\subsubsection{German}
The data was collected from YouTube using 2024 ICD-10 codes for diversity. We searched for videos related to diseases linked to these codes, but found limited human-labeled subtitles. To address this, we included German medical terms such as ``Krankenhaus'' and ``Krankheit'' in our searches. All selected videos had manually annotated captions with accurate timestamps. We prioritized videos longer than 20 minutes, then shorter ones, ensuring diversity in speakers by gender, accent, and context (lectures, discussions, interviews).


\subsection{Details of Data Quality Control}
\label{sec:details_data_quality_control}
The French transcript was triple-validated by a native French Literature lecturer and a C1-level linguist - a professional medical expert, ensuring transcription accuracy and alignment with the CC timing. The Chinese transcript was similarly validated by a native speaker and an HSK-5 level linguist  professional medical expert. The English transcript was initial reviewed by a TESOL-certified linguist, followed by cross-checking by three C1-level speakers, one of whom is a professional biomedical expert. Due to labor constraints, the German transcript was double-validated by a single C1-level professional biomedical expert.

All of our annotators were instructed to adhere to the following quality control procedures:
\begin{enumerate}
    \item Listen carefully to the audio recordings 
    \item Validate the human-annotated transcripts provided by professional YouTube channels by correcting minor inaccuracies or excluding transcripts deemed too erroneous
    \item Identify the start and end points of individual utterances
    \item Identify the speaker, recording conditions, accents, speaking roles (when applicable)
\end{enumerate}

\subsection{Data Processing}
\label{sec:data_processing}
Transcription errors often arise from time-stamp mismatches when segmenting long-form audio into shorter segments. Annotators use long-form audio to improve efficiency and capture extended contexts, such as discussions or lectures. Due to GPU memory limitations, training is restricted to short-form audio to prevent out-of-memory (OOM) issues. As a result, annotators split long transcripts, causing time-stamp mismatches, typically within one second. This can lead to missing words at the start or end of recordings, highlighting the limitations of human-labeled datasets, where annotators struggle to capture words occurring faster than one second \cite{wargo1967human}. For a standard conversational spontaneous ASR English dataset such as Switchboard \cite{godfrey1992switchboard}, the Word Error Rate (WER) for human annotators ranges from 5\% to 15\% \cite{stolcke2017comparing}. In contrast, for a more challenging real-world ASR dataset, the WER for human annotators without ASR support ranges from 17\% to 31\% \cite{mulholland2016comparison}.

In contrast, forced alignment can address this issue as machines can "listen" to words in 10ms-20ms intervals. However, forced alignment is limited by the quality of human-provided training data, making no transcript entirely accurate. To achieve "more perfect" transcripts, we employ a human-machine collaboration approach.

\begin{figure*}[h]
\centering
\includegraphics[width=\textwidth]{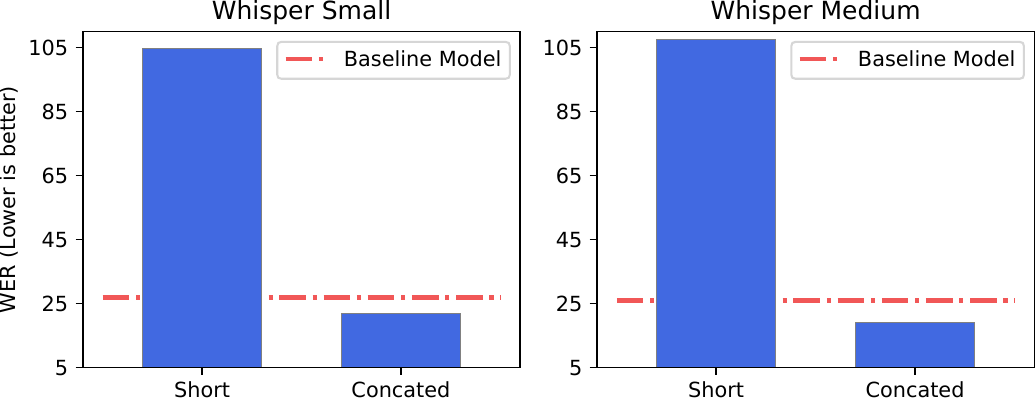}
\caption{Illustrating the performance comparison of Whisper models trained on two distinct audio segmentation approaches for German language data: human-segmented short audio clips and concatenated continual audio segments of approximately maximum 15 seconds in length. We evaluate performance using both Whisper Small and Whisper Medium model sizes. The results demonstrate a notable improvement in model performance when trained on concatenated audio, highlighting the efficacy of this data preparation technique in enhancing transcription accuracy in the German language context}
\label{fig:compared_concated}
\end{figure*}

To maximize the data quality for the training model. We implemented a tailored data quality control pipeline designed to address specific challenges inherent to multi-audio sources. The transcription process is often manual and can be inaccurate. Dividing audio into very short segments (i.e., less than 5 seconds) frequently results in serious misalignment with the transcriptions, which harms the training process. By concatenating these short segments, we created longer and more coherent training samples. This mitigates the misalignment problem and provides the model with a richer understanding of the patterns and intonation of spoken language. The results of the analysis before and after concatenation are shown in Figure \ref{fig:compared_concated}.

Additionally, extraneous noise text elements such as silence markers, filled pauses, and HTML tags, while present in raw transcripts, do not contribute to meaningful model learning. We removed these elements to focus the model's attention on relevant speech content. In particular, we chose to retain punctuation marks during the cleaning process. Punctuation plays a crucial role in conveying the nuances of spoken language, and its presence in training data encourages the model to generate transcripts that are not only accurate but also expressive and natural-sounding. 

\twocolumn
\subsection{Full Data Statistics}
\label{sec:full_data_stats}
\textit{MultiMed} dataset consists of multilingual audio recordings in five languages: Vietnamese, English, French, German, and Mandarin Chinese. Each audio clip was segmented into short snippets with an average length of approximately 6 seconds for Vietnamese and around 12 seconds for the other languages. This segmentation facilitated efficient training and improved the model's responsiveness to shorter speech segments. The dataset was subsequently uploaded to the Hugging Face platform for further training and analysis. The statistics of our data samples are described in Table \ref{table:dataset_statistics}.

Table \ref{table:data_literature} shows the statistics of the data set compared to all existing publicly available medical ASR datasets, based on our best knowledge of the current literature. As shown in the table, our \textit{MultiMed} dataset stands as the world's largest dataset in terms of total duration (150 hours of recordings), number of recording conditions (10), number of accents (16) and number of speaking roles (6).

\begin{table*}[t]
\centering
\begin{adjustbox}{width=\textwidth}
\begin{tabular}{l|c|ccccccc}
\hline
\textbf{Dataset} & \textbf{Venue} & \textbf{Dur.} & \textbf{Language} & \textbf{Nature} & \textbf{\#Rec. Cond.} & \textbf{\#Spk} & \textbf{\#Acc} & \textbf{\#Roles} \\ \hline
MultiMed$^1$ (ours) & - & 150h & Multiling. & Real-world & 10 & 198 & 16 & 6 \\
VietMed \cite{vietmed_dataset} & LREC-COLING & 16h & Vietnamese & Real-world & 8 & 61 & 6 & 6 \\
PriMock57$^2$ \cite{korfiatis2022primock57} & ACL & 9h & English & Simulated & 1 & 64 & 4 & 2 \\
Work by \citet{fareez2022dataset}$^3$ & Nature & 55h & English & Simulated & 1 & N/A & 1 & 2 \\
AfriSpeech-200$^4$ \cite{olatunji2023afrispeech} & TACL & $\approx$123h & African English & Read speech & 1 & N/A & N/A & 1 \\
myMediCon$^5$ \cite{htun2024mymedicon} & LREC-COLING & 11h & Burmese & Read speech & 1 & 12 & 5 & 2 \\ \hline
\end{tabular}
\end{adjustbox}
\caption{Dataset statistics in comparison with all existing works from left to right: Total duration in hours (h), language, nature of speech, number of recording conditions, number of speakers, number of accents, speaking roles.\\
$^1$In our dataset, only the number of recording conditions, speakers, accents and speaking roles for Vietnamese and English are identified because of technical and privacy issues. Therefore, the exact number of speakers and accents must be much larger than the currently reported number. 10 recording conditions include: Documentary, Interview, Lecture, News, Podcast, Webinar, Speech, Talk, Vlog, Workshop. 10 English accents include: Main US, Southern US, UK, Australian, Indian, Mexican, European, Japanese, Uzbekistan, Russian. 6 Vietnamese accents include: North, South Central Coast, South East, South West, Central Highland, North Central Coast.\\
$^2$Speech collected by simulated medical conversations between 2 speaking roles - clinicians and actors/actresses. 4 English accents include:  British English, European, other English, and other non-English.\\
$^3$Speech was recorded as patient-physician interviews (counted as 1 recording condition and 2 speaking roles) by West England speakers (counted as 1 accent)\\
$^4$AfriSpeech-200 dataset is a mix of general-domain and medical-domain speech. To our best understanding of the paper, we estimate the total duration of medical-domain speech to be around 123 hours. Recordings were collected by crowd-sourced workers to read aloud the medical transcripts (also known as read speech), thus both the number of recording conditions and speaking roles are counted as 1.\\
$^5$myMediCon dataset hired speakers to read aloud the translated medical transcripts from English corpus (thus known as read speech). 5 speakers' accents include: Native Burmese, Pa’O, Kachin, Dawei, and Mon. 2 speaking roles are patients and doctors.
}
\label{table:data_literature}
\end{table*}


\twocolumn
\section{Attention Encoder Decoder (AED)}
\label{sec:AED}
An ASR model aims to convert an audio signal into text by mapping an audio signal $x^{T}_{1} := x_{1}, x_{2}, ..., x_{T}$ of length $T$ to the most likely word sequence $w^{N}_{1}$ of length $N$. The word sequence probability is defined as:

\begin{equation}
\label{eq:word_sequence_general}
p(w_{1}^{N}|x_{1}^{T}) = \prod_{n=1}^{N} p(w_n|w_{1}^{n-1},x_{1}^{T}).    
\end{equation}

In the encoder-decoder architecture, given $D$ as the dimension size of the feature, the input audio signal matrix could be described as $x^{T}_{1} \in \mathbb{R}^{T \times D_{input}}$. For the sake of simplicity, downsampling prior to or within the encoder, achieved by a fixed factor, such as striding in a Convolutional Neural Network (CNN) is omitted. Consequently, the encoder output sequence is as follows:

\begin{equation}
h_{1}^{T} = Encoder(x_{1}^{T}) \in \mathbb{R}^{T \times D_{encoder}}. 
\end{equation}

Using a stack of Transformer (\scalebox{1.44}{$\tau$}) blocks \cite{vaswani2017attention}, the encoder output sequence is described as function composition:
\begin{equation}
h_{1}^{T} = {\scalebox{1.44}{$\tau$}}_{0} \circ ... \circ {\scalebox{1.44}{$\tau$}}_{N_{EncLayers}}(x_{1}^{T}).
\end{equation}

In the decoder, the probability for every single word is described as:

\begin{equation}
\begin{split}
p(w_n|w_{1}^{n-1},x_{1}^{T}) &= p(w_n|w_{1}^{n-1},h_{1}^{T}(x_{1}^{T}))\\
&= p(w_n|w_{1}^{n-1},h_{1}^{T}). 
\end{split}   
\end{equation}

Based on Eq. \ref{eq:word_sequence_general}, the word sequence probability given the output of encoder is formulated as:
\begin{equation}
p(w_{1}^{N}|x_{1}^{T}) = \prod_{n=1}^{N} p(w_n|w_{1}^{n-1},h_{1}^{T}).    
\end{equation}

Decoder hidden state is formulated as:
\begin{equation}
g_n = f(g_{n-1},w_{n-1}, c_n) \in \mathbb{R}^{D_{g}},
\end{equation}
where $f$ is neural network; $D_{g}$ is hidden state dimension; and $c_n$ is context vector, e.g. weighted sum of encoder outputs via attention mechanism.

The attention mechanism in the decoder is described by 3 components: context vector $c_n$, attention weights $\alpha_{n,t}$, and attention energy $e_{n,t}$:

\begin{equation}
\begin{split}
c_n &= \sum_{t=1}^{T} \alpha_{n,t} {h}_{t} \in \mathbb{R}^{D_{encoder}},\\
\alpha_{n,t} &= \frac{\exp(e_{n,t})}{\sum_{t'=1}^{T}\exp(e_{n,t'})} \\
&= Softmax_{T}(\exp(e_{n,t})) \in \mathbb{R},\\
e_{n,t} &= Align(g_{n-1}, h_t) \in \mathbb{R} \\
&= W_{2} \cdot \tanh(W_{1} \cdot [g_{n-1}, h_t]),
\end{split}
\end{equation}
where $n$ is decoder step; $t$ is encoder frame; $\alpha \in \mathbb{R}^{T \times N}$ is attention weight matrix; $\alpha_n \in \mathbb{R}^{T}$ is normalized probability distribution over $t$; $Softmax_{T}$ is Softmax function over spatial dimension $T$, not feature dimension; $W_{1} \in \mathbb{R}^{(D_{g}+D_{encoder}) \times D_{key}}$; $W_{2} \in \mathbb{R}^{D_{key}}$.

During decoding, the output probability distribution over vocabulary is described as:
\begin{equation}
\begin{split}
&p(w_{n} = *|w_{1}^{n-1}, h_{1}^{T})\\
&= Softmax(MLP(w_{n-1}, g_n, c_n)) \in \mathbb{R}^{N}, 
\end{split}
\end{equation}
where $MLP$ is Multi-layer Perceptron.

For training AED model,  sequence-level cross-entropy loss is employed:
\begin{equation}
\begin{split}
\mathscr{L}_{AED} &= - \sum_{(x_{1}^{T}, w_{1}^{N})} \log p(w_{1}^{N}|x_{1}^{T})\\
&= - \sum_{(x_{1}^{T}, w_{1}^{N})} \sum_{n=1}^{N} \log p(w_n|w_{1}^{n-1},x_{1}^{T}).
\end{split}
\end{equation}

In beam search process, the auxilary quantity for each unknown partial string (tree of partial hypotheses) $w_{1}^{n}$ is described as:
\begin{equation}
\begin{split}
Q(n; w_{1}^{n}) :&= \prod_{n'=1}^{n} p(w_{n'}|w_{0}^{n'-1},x_{1}^{T})\\
&= p(w_{n}|w_{0}^{n-1},x_{1}^{T}) \cdot Q(n-1, w_{1}^{n-1}).
\end{split}
\end{equation}

After eliminating the less likely hypotheses in the beam search process, the word sequence probability is determined by the most optimal hypothesis:
\begin{equation}
p(w_{1}^{N}|x_{1}^{T}) = Q(N; w_{1}^{N}).    
\end{equation}

\twocolumn
\section{Full Experimental Setups}
\subsection{Hyperparameter Tuning}
\label{sec:hyperparam_tuning}
The training process leveraged powerful A100 SXM4 GPUs. To ensure consistent results, we fixed the random seed at 42 throughout the training runs. For all models, we adopted a common training configuration with a batch size of 8, a learning rate of 0.0001, and 20 training epochs. We applied several data pre-processing techniques during training, including lowercasing text, removing punctuation, and normalizing the audio input. In addition, we trained each model on a language-specific subset of the dataset to optimize its performance for the targeted language.

The optimizer chosen for training was Adam \cite{kingma2014adam} with the standard betas configuration (0.9, 0.999) and an epsilon value of 1e-8. We employed a linear learning rate scheduler with a warmup period of 100 steps to gradually increase the learning rate during the initial training phase. No data augmentation such as SpecAugment \cite{park19e_interspeech} was applied.

\subsection{Details of Evaluation Metrics}
\label{sec:details_eval_metrics}
To assess the performance of the ASR models, we used two standard evaluation metrics: WER and CER. Lower WER and CER scores indicate better model performance in terms of accurately transcribing spoken audio.

WER focuses on the accuracy of recognized words. It calculates the percentage of errors made at the word level, including insertions, deletions, and substitutions compared to the ground truth reference transcript, as described in Equation \ref{eq.wer}.

\begin{equation} 
\label{eq.wer}
WER = \frac{S + D + I}{N} = \frac{S + D + I}{S + D + C}
\end{equation}
where $S$ is the number of word substitutions, $D$ is the number of word deletions, $I$ is the number of word insertions, $C$ is the number of correct words, and $N$ is the number of words in the reference data $(N=S+D+C)$.

Generally speaking, $S$ represents the count of replaced words, $D$ denotes the count of omitted words present in the reference data but absent in the ASR hypothesis, and $I$ indicates the count of inserted words present in the ASR hypothesis but absent in the reference data. The alignment process between the ASR hypothesis and the reference data proceeds sequentially from left to right.

WER measures the number of insertions, deletions, and substitutions made at the word level, while the CER focuses on errors at the character level, as illustrated in Equation \ref{eq.cer}. 

\begin{equation} 
\label{eq.cer}
CER = \frac{S_c + D_c + I_c}{N_c} = \frac{S_c + D_c + I_c}{S_c + D_c + C_c}
\end{equation}
where $S_c$ is the number of character substitutions, $D_c$ is the number of character deletions, $I_c$ is the number of character insertions, $C_c$ is the number of correct characters, and $N_c$ is the number of characters in the reference data $(N_c=S_c+D_c+C_c)$.

\section{Details of Hybrid ASR Experiments}
\label{sec.extra_experiments_AED_Hybrid}
\subsection{Hybrid wav2vec 2.0}
\label{sec:hybrid_wav2vec2}
\subsubsection{Hybrid ASR}
An ASR model aims to convert an audio signal into text by mapping an audio signal $x^{T}_{1}$ of length $T$ to the most likely word sequence $w^{N}_{1}$ of length $N$. 
The relation $w^{*}$ between the acoustic and word sequence is:
\begin{equation}
w^{*} = \operatorname{arg}\max_{w_1^N} \, p(w_{1}^{N}|x_{1}^{T})    
\end{equation}

\textbf{Bayes theorem}: By applying Bayes’ Theorem, the probability $p(x)$ can be ignored during the maximization process, as it functions only as a normalization constant and does not influence the final result.

\begin{equation}
    \begin{split}
     p(w_{1}^{N}|x_{1}^{T}) &= \frac{p(x_{1}^{T}|w_{1}^{N})p(w_{1}^{N})}{p(x_{1}^{T})} \\
    & \propto p(x_{1}^{T}|w_{1}^{N})p(w_{1}^{N})    
    \end{split}
\end{equation}
Therefore:
\begin{equation}
w^{*} = \operatorname{arg}\max_{w_1^N}  \underbrace{p(x_{1}^{T}|w_{1}^{N})}_{\text{acoustic model}}\cdot\underbrace{p(w_{1}^{N})}_{\text{language model}}
\end{equation}

\textbf{Acoustic modeling}: First, alignments between the acoustic observations $x_1^T$ and labels $w_1^N$ are obtained by using Gaussian-Mixture-Model/Hidden-Markov-Model (GMM/HMM) as labels for Deep-Neural-Network/Hidden-Markov-Model (DNN/HMM) training (DNN is wav2vec 2.0 encoder \cite{baevski2020wav2vec} in this case).

\begin{equation}
\begin{split}
&p(x_1^T|w_1^N) = \sum_{[s_1^T]}\prod_{t=1}^Tp(x_t, s_t|s_{t-1}, w_1^N) \\
&= \sum_{[s_1^T]}\prod_{t=1}^T\underbrace{p(s_t|s_{t-1}, w_1^N)}_{\text{transition prob.}}\cdot \underbrace{p(x_t|s_t, s_{t-1}, w_1^N)}_{\text{emission prob.}}    
\end{split}
\end{equation}

\textbf{GMM/HMM}: The labels used in the acoustic modeling are context-dependent phonemes (triphones), instead of BPE subword units like in AED. During the GMM/HMM process, a CART (Classification and Regression Tree) \cite{breiman2017classification} is used to link the states $s$. The GMM is a weighted sum over $K$ normal distributions and is calculated as:

\begin{align}
	p(x_t|s_t, s_{t-1}, w_1^N) = \sum_{i=1}^K c_i \cdot \mathcal{N}(x_t|\mu_i, \sigma_i^2),
\end{align}

resulting in a multimodal emission probability with parameters $\mu_{i}, \sigma_{i}$ and mixture weights $c_i$ for $i\in\llbracket1,K\rrbracket$. The mixture weights are non-negative and sum up to unity.

\textbf{DNN/HMM}: The posterior probability $p(a_{s_t}|x_1^T)$ could be discriminatively modeled using DNN (wav2vec 2.0 encoder), resulting in the DNN/HMM approach. The emission probability in the HMM could be calculated using the Bayes rule:

\begin{align}
	p(x_1^T|a_{s_t}) = \frac{p(a_{s_t}|x_1^T)p(x_1^T)}{p(a_{s_t})}.
\end{align}

The probability $p(a_{s_t})$ could be estimated as the relative frequency of $a_{s_t}$. For a simplified Bayes decision rule, the probability $p(x_1^T)$ is removed.

\textbf{Decoding}: 
During the ASR decoding process, the acoustic model and n-gram language model \cite{ney1994structuring} should be combined based on the Bayes decision rule using Viterbi decoding algorithm \cite{viterbi} which recursively calculates the maximum path to a find best-path in the alignment graph of all possible predicted words to the acoustic observations:
\begin{equation}
\begin{split}
w_1^N &= \operatorname{arg}\max_{N,w_1^N}p\Bigl(\prod_{n=1}^Np(w_n|w_{n-m}^{n-1}) \\
&\cdot \max_{[s_1^T]}\prod_{t=1}^Tp(x_t,s_t|s_{t-1}, w_1^N)\Bigr)    
\end{split}
\end{equation}
Afterwards, beam search (acoustic model and n-gram language model pruning) is employed to solely focus on the most promising predicted words at each time step $t$ \cite{ortmanns1997word}.

\subsubsection{Modified wav2vec 2.0}
The model consists of a multi-layer convolutional neural network feature extractor $CNN$ that receives $T$ time-step raw audio waveform $x^{T}_{1} := x_{1}, x_{2}, ..., x_{T}$ (or $x$ for simplification, $x \in R^{T \times 1}$) as input and produces latent speech representations $x^{FE} \in R^{T \times 1}$.
These representations are then pushed into a stack of Transformer (\scalebox{1.44}{$\tau$}) layers \cite{vaswani2017attention} which generates contextualized representations for Softmax $SM$ classification.

In the scope of this work, we mathematically formulate our modified architecture for Hybrid wav2vec 2.0 as follows.

\textbf{Wave normalization\footnote{Our modification of wav2vec 2.0 architecture for Hybrid ASR}}: The raw audio waveform $x$ is first normalized to the range between 0 and 1 by the wave normalization layer $WaveNorm$ before being pushed into the feature extractor, as shown in Equation \ref{Eq:WaveNorm}.
\begin{equation}
\label{Eq:WaveNorm}
\begin{split}
x^{WaveNorm} &= WaveNorm(x)\\
&=LayerNorm(x) \in R^{T \times 1}
\end{split}
\end{equation}
$WaveNorm$ could be either layer normalization $LayerNorm$ \cite{LayerNorm} or batch normalization $BatchNorm$ \cite{BatchNorm}. 

\textbf{Feature extractor}: The normalized raw audio waveform is pushed into a stack of CNN layers and a feed-forward (FFW) layer.
\begin{equation}
\begin{split}
x^{FE} &= FeatureExtractor(x)\\
       &= FFN \circ CNNs \circ WaveNorm(x)
       \end{split}
\end{equation}

\textbf{Time-downsampling in feature extractor\footnote{Our modification of wav2vec 2.0 architecture for sampling rate mismatch between pre-trained models and fine-tuned dataset}}: When there is a sampling rate mismatch, the feature extractor of 16 kHz pre-trained models can be modified to handle 8 kHz sampled data while still producing representations with the same 20 ms frame shift. By halving the stride of a convolutional layer in a stack of CNN layers in the feature extractor, we will receive features at the desired frame rate while reducing the downsampling factor from the waveform to the feature frames by a factor of 2.

\begin{equation}
\begin{split}
&x^{FE} := TimeDownsampling(x^{FE}) |\\
&x^{FE} \in R^{\frac{1}{2}T_{FE} \times F_{FE}}
\end{split}
\end{equation}

In a generalized formulation shown in Equation \ref{Eq:Generalized_Time-downsampling}, the time-downsampling could be done given a general time-downsampling factor $TDF$
\begin{equation}
\label{Eq:Generalized_Time-downsampling}
\begin{split}
&x^{FE} := TimeDownsampling(x^{FE}) |\\
&x^{FE} \in R^{\frac{1}{TDF}T_{FE} \times F_{FE}}
\end{split}
\end{equation}

\textbf{Transformer as contextualized encoder}: In an arbitrary $l$-th transformer layer, the output $x^{\tau}_{l}$ is briefly defined as:

\begin{equation}
\begin{split}
x^{\tau}_{l} &= \scalebox{1.44}{$\tau$}(x^{\tau}_{l-1})\\
&=FFW \circ MHA(x^{\tau}_{l-1})
\end{split}
\end{equation}
where $MHA$ is multi-head attention which is a function defined by self-attention functions $SA$:

\begin{equation}
MHA(x^{\tau}_{l-1}) = SA(x^{\tau}_{l-1}) + x^{\tau}_{l-1}
\end{equation}

Then, we have a full equation for an arbitary $l$-th Transformer layer:
\begin{equation}
\begin{split}
x^{\tau}_{l} &= FFW(MHA(x^{\tau}_{l-1})) + MHA(x^{\tau}_{l-1}) \\
&= FFW(SA(x^{\tau}_{l-1}) + x^{\tau}_{l-1})\\
&+ \left [SA(x^{\tau}_{l-1}) + x^{\tau}_{l-1}\right ]
\end{split}
\end{equation}

For layer-wise formulation, the 0-th Transformer layer (the first layer) is connected to the feature extractor, which is defined as:
\begin{equation}
x^{\tau}_{0} = \scalebox{1.44}{$\tau$}(x^{FE})
\end{equation}

Given an $L$-Transformer-layer wav2vec 2.0 architecture, the $L-1$-th Transformer layer (the final layer) is defined as a chain function as:
\begin{equation}
\begin{split}
x^{\tau}_{L-1} &= \scalebox{1.44}{$\tau$}(x^{\tau}_{L-2}) \\
&= \scalebox{1.44}{$\tau$} \circ \scalebox{1.44}{$\tau$} \circ ... \circ \scalebox{1.44}{$\tau$}(x^{\tau}_{0})\\
&= \scalebox{1.44}{$\tau$} \circ \scalebox{1.44}{$\tau$} \circ ... \circ \scalebox{1.44}{$\tau$} \circ \scalebox{1.44}{$\tau$}(x^{FE})
\end{split}
\end{equation}
where $L$ is the total number of Transformer layers in the encoder, layer indices start from $0$ to $L-1$.

\textbf{Time-reupsampling}: For wav2vec 2.0 architecture, regardless of whether a sampling rate mismatch exists or not, it is necessary to re-upsample the final Transformer layer prior to its input into a Softmax layer for frame-wise classification. Failure to do so would lead to a discrepancy in the number of time frames during the calculation of the frame-wise loss objective function. Consequently, a FFW necessitates upsampling to ensure alignment with the rest of the architecture.
\begin{equation}
\begin{split}
x_{Reup} &:= TimeReupsampling(x^{\tau}_{L-1})\\
&:= FFW(x^{\tau}_{L-1})|\\
&x_{Reup} \in R^{T \times d}
\end{split}
\end{equation}
where $d$ is the size of context-dependent states (CDS), or size of CART labels. 

\textbf{Hypothesis (output)}: Finally, $x_{Reup}$ goes to a Softmax layer $SM$ to produce a matrix of hypotheses $z \in R^{T \times d}$.
\begin{equation}
\begin{split}
z := SM(x_{Reup})| z \in R^{T \times d}
\end{split}
\end{equation}

\textbf{Loss function}: The hypothesis matrix $z$ is compared with the ground truth $y$ to calculate the frame-wise cross-entropy (CE) loss matrix $\mathscr{L}(z,y) \in R^{T \times d}$. The total loss value is the sum of all the elements in the loss matrix $\mathscr{L}(z,y)$. 
\begin{equation}
\begin{split}
\mathscr{L}(z,y) &:= \mathscr{L}_{f}(z,y) = { \left \| \mathscr{L}(z,y) \right \| }^{1}\\
&:= -[y \cdot \log(z)], \quad f=CE\\
&>0 \quad \forall \log \in \{\log_{2}, \log_{n}, \log_{10}\}
\end{split}
\end{equation}

\subsection{Experimental Setups}
\begin{table*}[h]
\centering
\begin{tabular}{|cc|cc|cc|cc|}
\hline
\multicolumn{2}{|c|}{\textbf{Trained lexicon}} & \multicolumn{2}{c|}{\textbf{Language model}} & \multicolumn{2}{c|}{\textit{\textbf{dev}}} & \multicolumn{2}{c|}{\textit{\textbf{test}}} \\ \hline
\multicolumn{1}{|c|}{\textbf{\#words}} & \textbf{\#vocab} & \multicolumn{1}{c|}{\textbf{\#words}} & \textbf{Size (in MBs)} & \multicolumn{1}{c|}{\textbf{OOV}} & \textbf{PPL} & \multicolumn{1}{c|}{\textbf{OOV}} & \textbf{PPL} \\ \hline
\multicolumn{1}{|c|}{17,000} & 5295 & \multicolumn{1}{c|}{8.5M} & 98 & \multicolumn{1}{c|}{0.76\%} & 66 & \multicolumn{1}{c|}{0.66\%} & 84 \\ \hline
\end{tabular}%
\caption{Statistics of 4-gram language model and augmented lexicon for hybrid ASR training, including for both GMM-HMM and wav2vec 2.0 training. OOVs and Perplexities (PPLs) are reported on our Vietnamese dev and test set.}
\label{tab.LM_lexicon_results}
\end{table*}
For n-gram language modelling and the initialization of GMM-HMM, we used the same configurations and hyperparameters as in \cite{luescher2022:hykist}. We employed the BABEL project's seed lexicon and augmented it with additional Vietnamese text data. Using the toolkit Sequitur Grapheme-To-Phoneme\footnote{https://github.com/sequitur-g2p/sequitur-g2p} \cite{G2P_toolkit} - the conversion tool on pronunciation lexicon, the seed lexicon from BABEL was extended, creating the augmented lexicon for training. The statistics for the n-gram language model and the augmented lexicon are shown in Table \ref{tab.LM_lexicon_results}.

The labels for the acoustic model were generalized triphone states obtained by CART with 4501 labels. During GMM-HMM process, we found that WERs on the Vietnamese test sets of Speaker Adaptive Training (SAT) was quite comparable to Speaker Adaptive Training + Vocal Tract Length Normalization (SAT+VTLN). So, we fed SAT alignments into wav2vec 2.0 as input for the Hybrid ASR training.

For self-supervised wav2vec 2.0 training \cite{facebook2020wav2vec2} and fine-tuning, we used the same vanilla setups and hyperparameters in \cite{bachelorthesis}. All models had 123M parameters including 7 CNN layers and 8 Transformer layers, as shown in Table \ref{tab.AED_vs_Hybrid} in the main paper. The last CNN layer had a stride halved for adaptation to the 8kHz data. The pre-training epoch that led to the best WERs on dev was used to fine-tune with framewise CE loss. The SpecAugment \cite{park2019specaugment} was employed during 33 fine-tuning epochs.

We employed RETURNN framework \cite{zeyer2018returnn} for supervised training (fine-tuning the wav2vec 2.0 models) and Fairseq \cite{facebook2019fairseq} for self-supervised wav2vec 2.0 training on the unlabeled data. ASR decoding was performed using the RASR toolkit \cite{rybach2011rasr}. The pre-trained wav2vec 2.0 models from Fairseq (in Pytorch) were converted to RETURNN models (in Tensorflow) with our PyTorch-to-RETURNN toolkit\footnote{https://github.com/rwth-i6/pytorch-to-returnn}.

\subsection{Extra Experimental Results}

Table \ref{tab.breakdown_HybridASR_perSpeaker} shows the breakdown per speaker in the Vietnamese test set of the Hybrid ASR results in Table \ref{tab.AED_vs_Hybrid}. Two pre-trained wav2vec 2.0 models were used for fine-tuning on the Vietnamese set: XLSR-53-Viet and w2v2-Viet, leading to WERs on test set 28.8\%, 29.0\% respectively.

\onecolumn
\begin{longtable}[c]{|lcccccccc|}
\hline
\multicolumn{1}{|c|}{\textbf{Speaker ID}} & \multicolumn{1}{c|}{\textbf{\# Snt}} & \multicolumn{1}{c|}{\textbf{\# Wrd}} & \multicolumn{1}{c|}{\textbf{Corr}} & \multicolumn{1}{c|}{\textbf{Sub}} & \multicolumn{1}{c|}{\textbf{Del}} & \multicolumn{1}{c|}{\textbf{Ins}} & \multicolumn{1}{c|}{\textbf{Err}} & \multicolumn{1}{c|}{\textbf{S.Err}} \\ \hline
\endfirsthead
\endhead
\multicolumn{9}{c}{\textbf{XLSR-53-Viet}} \\ \hline
\multicolumn{1}{|l|}{vietmed\_002} & \multicolumn{1}{c|}{363} & \multicolumn{1}{c|}{7631} & \multicolumn{1}{c|}{58.5} & \multicolumn{1}{c|}{30.9} & \multicolumn{1}{c|}{10.6} & \multicolumn{1}{c|}{6.6} & \multicolumn{1}{c|}{48.1} & \multicolumn{1}{c|}{100.0} \\ \hline
\multicolumn{1}{|l|}{vietmed\_004} & \multicolumn{1}{c|}{446} & \multicolumn{1}{c|}{10575} & \multicolumn{1}{c|}{68.3} & \multicolumn{1}{c|}{18.5} & \multicolumn{1}{c|}{13.2} & \multicolumn{1}{c|}{4.9} & \multicolumn{1}{c|}{36.6} & \multicolumn{1}{c|}{100.0} \\ \hline
\multicolumn{1}{|l|}{vietmed\_014\_a} & \multicolumn{1}{c|}{18} & \multicolumn{1}{c|}{491} & \multicolumn{1}{c|}{88.6} & \multicolumn{1}{c|}{3.1} & \multicolumn{1}{c|}{8.4} & \multicolumn{1}{c|}{5.9} & \multicolumn{1}{c|}{17.3} & \multicolumn{1}{c|}{100.0} \\ \hline
\multicolumn{1}{|l|}{vietmed\_014\_b} & \multicolumn{1}{c|}{164} & \multicolumn{1}{c|}{4034} & \multicolumn{1}{c|}{77.2} & \multicolumn{1}{c|}{11.8} & \multicolumn{1}{c|}{11.1} & \multicolumn{1}{c|}{3.7} & \multicolumn{1}{c|}{26.5} & \multicolumn{1}{c|}{100.0} \\ \hline
\multicolumn{1}{|l|}{vietmed\_015\_a} & \multicolumn{1}{c|}{73} & \multicolumn{1}{c|}{1779} & \multicolumn{1}{c|}{86.1} & \multicolumn{1}{c|}{5.5} & \multicolumn{1}{c|}{8.4} & \multicolumn{1}{c|}{3.9} & \multicolumn{1}{c|}{17.8} & \multicolumn{1}{c|}{98.6} \\ \hline
\multicolumn{1}{|l|}{vietmed\_015\_b} & \multicolumn{1}{c|}{297} & \multicolumn{1}{c|}{5669} & \multicolumn{1}{c|}{83.3} & \multicolumn{1}{c|}{6.9} & \multicolumn{1}{c|}{9.8} & \multicolumn{1}{c|}{4.2} & \multicolumn{1}{c|}{20.9} & \multicolumn{1}{c|}{96.6} \\ \hline
\multicolumn{1}{|l|}{vietmed\_015\_c} & \multicolumn{1}{c|}{55} & \multicolumn{1}{c|}{1010} & \multicolumn{1}{c|}{69.4} & \multicolumn{1}{c|}{14.4} & \multicolumn{1}{c|}{16.2} & \multicolumn{1}{c|}{5.5} & \multicolumn{1}{c|}{36.1} & \multicolumn{1}{c|}{100.0} \\ \hline
\multicolumn{1}{|l|}{vietmed\_017\_a} & \multicolumn{1}{c|}{47} & \multicolumn{1}{c|}{1104} & \multicolumn{1}{c|}{78.3} & \multicolumn{1}{c|}{12.0} & \multicolumn{1}{c|}{9.7} & \multicolumn{1}{c|}{4.6} & \multicolumn{1}{c|}{26.4} & \multicolumn{1}{c|}{100.0} \\ \hline
\multicolumn{1}{|l|}{vietmed\_017\_b} & \multicolumn{1}{c|}{86} & \multicolumn{1}{c|}{2061} & \multicolumn{1}{c|}{81.5} & \multicolumn{1}{c|}{9.8} & \multicolumn{1}{c|}{8.6} & \multicolumn{1}{c|}{5.0} & \multicolumn{1}{c|}{23.5} & \multicolumn{1}{c|}{100.0} \\ \hline
\multicolumn{1}{|l|}{vietmed\_018\_a} & \multicolumn{1}{c|}{63} & \multicolumn{1}{c|}{1527} & \multicolumn{1}{c|}{76.0} & \multicolumn{1}{c|}{11.9} & \multicolumn{1}{c|}{12.2} & \multicolumn{1}{c|}{19.4} & \multicolumn{1}{c|}{43.5} & \multicolumn{1}{c|}{100.0} \\ \hline
\multicolumn{1}{|l|}{vietmed\_018\_b} & \multicolumn{1}{c|}{192} & \multicolumn{1}{c|}{5293} & \multicolumn{1}{c|}{76.7} & \multicolumn{1}{c|}{10.8} & \multicolumn{1}{c|}{12.5} & \multicolumn{1}{c|}{6.9} & \multicolumn{1}{c|}{30.2} & \multicolumn{1}{c|}{100.0} \\ \hline
\multicolumn{1}{|l|}{vietmed\_018\_c} & \multicolumn{1}{c|}{118} & \multicolumn{1}{c|}{2761} & \multicolumn{1}{c|}{76.5} & \multicolumn{1}{c|}{10.9} & \multicolumn{1}{c|}{12.5} & \multicolumn{1}{c|}{8.2} & \multicolumn{1}{c|}{31.7} & \multicolumn{1}{c|}{100.0} \\ \hline
\multicolumn{1}{|l|}{vietmed\_018\_d} & \multicolumn{1}{c|}{20} & \multicolumn{1}{c|}{412} & \multicolumn{1}{c|}{55.1} & \multicolumn{1}{c|}{19.7} & \multicolumn{1}{c|}{25.2} & \multicolumn{1}{c|}{5.6} & \multicolumn{1}{c|}{50.5} & \multicolumn{1}{c|}{100.0} \\ \hline
\multicolumn{1}{|l|}{vietmed\_018\_e} & \multicolumn{1}{c|}{5} & \multicolumn{1}{c|}{76} & \multicolumn{1}{c|}{56.6} & \multicolumn{1}{c|}{19.7} & \multicolumn{1}{c|}{23.7} & \multicolumn{1}{c|}{7.9} & \multicolumn{1}{c|}{51.3} & \multicolumn{1}{c|}{100.0} \\ \hline
\multicolumn{1}{|l|}{vietmed\_018\_f} & \multicolumn{1}{c|}{25} & \multicolumn{1}{c|}{639} & \multicolumn{1}{c|}{64.8} & \multicolumn{1}{c|}{20.7} & \multicolumn{1}{c|}{14.6} & \multicolumn{1}{c|}{6.9} & \multicolumn{1}{c|}{42.1} & \multicolumn{1}{c|}{100.0} \\ \hline
\multicolumn{1}{|l|}{vietmed\_019\_a} & \multicolumn{1}{c|}{58} & \multicolumn{1}{c|}{1490} & \multicolumn{1}{c|}{77.7} & \multicolumn{1}{c|}{10.3} & \multicolumn{1}{c|}{12.0} & \multicolumn{1}{c|}{6.8} & \multicolumn{1}{c|}{29.1} & \multicolumn{1}{c|}{100.0} \\ \hline
\multicolumn{1}{|l|}{vietmed\_019\_b} & \multicolumn{1}{c|}{116} & \multicolumn{1}{c|}{2776} & \multicolumn{1}{c|}{77.5} & \multicolumn{1}{c|}{11.1} & \multicolumn{1}{c|}{11.4} & \multicolumn{1}{c|}{6.6} & \multicolumn{1}{c|}{29.1} & \multicolumn{1}{c|}{100.0} \\ \hline
\multicolumn{1}{|l|}{vietmed\_023} & \multicolumn{1}{c|}{390} & \multicolumn{1}{c|}{7414} & \multicolumn{1}{c|}{85.5} & \multicolumn{1}{c|}{9.1} & \multicolumn{1}{c|}{5.3} & \multicolumn{1}{c|}{4.6} & \multicolumn{1}{c|}{19.1} & \multicolumn{1}{c|}{97.7} \\ \hline
\multicolumn{1}{|l|}{vietmed\_024} & \multicolumn{1}{c|}{376} & \multicolumn{1}{c|}{7425} & \multicolumn{1}{c|}{86.6} & \multicolumn{1}{c|}{7.0} & \multicolumn{1}{c|}{6.4} & \multicolumn{1}{c|}{5.4} & \multicolumn{1}{c|}{18.9} & \multicolumn{1}{c|}{98.7} \\ \hline
\multicolumn{1}{|l|}{vietmed\_025\_a} & \multicolumn{1}{c|}{101} & \multicolumn{1}{c|}{2280} & \multicolumn{1}{c|}{80.8} & \multicolumn{1}{c|}{10.1} & \multicolumn{1}{c|}{9.1} & \multicolumn{1}{c|}{5.0} & \multicolumn{1}{c|}{24.2} & \multicolumn{1}{c|}{100.0} \\ \hline
\multicolumn{1}{|l|}{vietmed\_025\_b} & \multicolumn{1}{c|}{91} & \multicolumn{1}{c|}{1838} & \multicolumn{1}{c|}{82.5} & \multicolumn{1}{c|}{9.2} & \multicolumn{1}{c|}{8.3} & \multicolumn{1}{c|}{5.3} & \multicolumn{1}{c|}{22.8} & \multicolumn{1}{c|}{98.9} \\ \hline
\multicolumn{1}{|l|}{vietmed\_026} & \multicolumn{1}{c|}{21} & \multicolumn{1}{c|}{355} & \multicolumn{1}{c|}{55.8} & \multicolumn{1}{c|}{29.9} & \multicolumn{1}{c|}{14.4} & \multicolumn{1}{c|}{7.3} & \multicolumn{1}{c|}{51.5} & \multicolumn{1}{c|}{100.0} \\ \hline
\multicolumn{1}{|l|}{vietmed\_027\_a} & \multicolumn{1}{c|}{29} & \multicolumn{1}{c|}{710} & \multicolumn{1}{c|}{85.5} & \multicolumn{1}{c|}{6.5} & \multicolumn{1}{c|}{8.0} & \multicolumn{1}{c|}{5.2} & \multicolumn{1}{c|}{19.7} & \multicolumn{1}{c|}{100.0} \\ \hline
\multicolumn{1}{|l|}{vietmed\_027\_b} & \multicolumn{1}{c|}{64} & \multicolumn{1}{c|}{1454} & \multicolumn{1}{c|}{76.3} & \multicolumn{1}{c|}{14.6} & \multicolumn{1}{c|}{9.1} & \multicolumn{1}{c|}{6.2} & \multicolumn{1}{c|}{29.8} & \multicolumn{1}{c|}{98.4} \\ \hline
\multicolumn{1}{|l|}{vietmed\_028\_a} & \multicolumn{1}{c|}{106} & \multicolumn{1}{c|}{2617} & \multicolumn{1}{c|}{83.7} & \multicolumn{1}{c|}{8.7} & \multicolumn{1}{c|}{7.6} & \multicolumn{1}{c|}{4.6} & \multicolumn{1}{c|}{20.9} & \multicolumn{1}{c|}{99.1} \\ \hline
\multicolumn{1}{|l|}{vietmed\_028\_b} & \multicolumn{1}{c|}{21} & \multicolumn{1}{c|}{475} & \multicolumn{1}{c|}{77.7} & \multicolumn{1}{c|}{11.8} & \multicolumn{1}{c|}{10.5} & \multicolumn{1}{c|}{5.9} & \multicolumn{1}{c|}{28.2} & \multicolumn{1}{c|}{95.2} \\ \hline
\multicolumn{1}{|l|}{vietmed\_029} & \multicolumn{1}{c|}{92} & \multicolumn{1}{c|}{2240} & \multicolumn{1}{c|}{83.8} & \multicolumn{1}{c|}{7.9} & \multicolumn{1}{c|}{8.3} & \multicolumn{1}{c|}{5.3} & \multicolumn{1}{c|}{21.6} & \multicolumn{1}{c|}{100.0} \\ \hline
\multicolumn{1}{|l|}{\textbf{Sum/Avg}} & \multicolumn{1}{c|}{3437} & \multicolumn{1}{c|}{76136} & \multicolumn{1}{c|}{76.9} & \multicolumn{1}{c|}{13.0} & \multicolumn{1}{c|}{10.1} & \multicolumn{1}{c|}{5.7} & \multicolumn{1}{c|}{\textbf{28.8}} & \multicolumn{1}{c|}{99.2} \\ \hline
\multicolumn{1}{|l|}{\textbf{Mean}} & \multicolumn{1}{c|}{127.3} & \multicolumn{1}{c|}{2819.9} & \multicolumn{1}{c|}{75.9} & \multicolumn{1}{c|}{12.7} & \multicolumn{1}{c|}{11.4} & \multicolumn{1}{c|}{6.2} & \multicolumn{1}{c|}{30.3} & \multicolumn{1}{c|}{99.4} \\ \hline
\multicolumn{1}{|l|}{\textbf{S.D.}} & \multicolumn{1}{c|}{129.6} & \multicolumn{1}{c|}{2743.3} & \multicolumn{1}{c|}{10.0} & \multicolumn{1}{c|}{6.7} & \multicolumn{1}{c|}{4.6} & \multicolumn{1}{c|}{2.9} & \multicolumn{1}{c|}{11.0} & \multicolumn{1}{c|}{1.2} \\ \hline
\multicolumn{1}{|l|}{\textbf{Median}} & \multicolumn{1}{c|}{86.0} & \multicolumn{1}{c|}{1838.0} & \multicolumn{1}{c|}{77.7} & \multicolumn{1}{c|}{10.9} & \multicolumn{1}{c|}{10.5} & \multicolumn{1}{c|}{5.5} & \multicolumn{1}{c|}{28.2} & \multicolumn{1}{c|}{100.0} \\ \hline
\multicolumn{9}{c}{\textbf{w2v2-Viet}} \\ \hline
\multicolumn{1}{|l|}{vietmed\_002} & \multicolumn{1}{c|}{363} & \multicolumn{1}{c|}{7631} & \multicolumn{1}{c|}{56.6} & \multicolumn{1}{c|}{31.0} & \multicolumn{1}{c|}{12.4} & \multicolumn{1}{c|}{6.1} & \multicolumn{1}{c|}{49.5} & \multicolumn{1}{c|}{100.0} \\ \hline
\multicolumn{1}{|l|}{vietmed\_004} & \multicolumn{1}{c|}{446} & \multicolumn{1}{c|}{10575} & \multicolumn{1}{c|}{65.5} & \multicolumn{1}{c|}{20.6} & \multicolumn{1}{c|}{13.9} & \multicolumn{1}{c|}{4.5} & \multicolumn{1}{c|}{39.0} & \multicolumn{1}{c|}{99.6} \\ \hline
\multicolumn{1}{|l|}{vietmed\_014\_a} & \multicolumn{1}{c|}{18} & \multicolumn{1}{c|}{491} & \multicolumn{1}{c|}{89.0} & \multicolumn{1}{c|}{2.9} & \multicolumn{1}{c|}{8.1} & \multicolumn{1}{c|}{6.1} & \multicolumn{1}{c|}{17.1} & \multicolumn{1}{c|}{100.0} \\ \hline
\multicolumn{1}{|l|}{vietmed\_014\_b} & \multicolumn{1}{c|}{164} & \multicolumn{1}{c|}{4034} & \multicolumn{1}{c|}{77.6} & \multicolumn{1}{c|}{12.7} & \multicolumn{1}{c|}{9.7} & \multicolumn{1}{c|}{4.9} & \multicolumn{1}{c|}{27.3} & \multicolumn{1}{c|}{100.0} \\ \hline
\multicolumn{1}{|l|}{vietmed\_015\_a} & \multicolumn{1}{c|}{73} & \multicolumn{1}{c|}{1779} & \multicolumn{1}{c|}{87.5} & \multicolumn{1}{c|}{5.0} & \multicolumn{1}{c|}{7.5} & \multicolumn{1}{c|}{3.7} & \multicolumn{1}{c|}{16.1} & \multicolumn{1}{c|}{98.6} \\ \hline
\multicolumn{1}{|l|}{vietmed\_015\_b} & \multicolumn{1}{c|}{297} & \multicolumn{1}{c|}{5669} & \multicolumn{1}{c|}{83.3} & \multicolumn{1}{c|}{6.3} & \multicolumn{1}{c|}{10.4} & \multicolumn{1}{c|}{3.7} & \multicolumn{1}{c|}{20.3} & \multicolumn{1}{c|}{96.6} \\ \hline
\multicolumn{1}{|l|}{vietmed\_015\_c} & \multicolumn{1}{c|}{55} & \multicolumn{1}{c|}{1010} & \multicolumn{1}{c|}{68.6} & \multicolumn{1}{c|}{13.8} & \multicolumn{1}{c|}{17.6} & \multicolumn{1}{c|}{4.6} & \multicolumn{1}{c|}{35.9} & \multicolumn{1}{c|}{100.0} \\ \hline
\multicolumn{1}{|l|}{vietmed\_017\_a} & \multicolumn{1}{c|}{47} & \multicolumn{1}{c|}{1104} & \multicolumn{1}{c|}{78.4} & \multicolumn{1}{c|}{12.0} & \multicolumn{1}{c|}{9.5} & \multicolumn{1}{c|}{4.7} & \multicolumn{1}{c|}{26.3} & \multicolumn{1}{c|}{100.0} \\ \hline
\multicolumn{1}{|l|}{vietmed\_017\_b} & \multicolumn{1}{c|}{86} & \multicolumn{1}{c|}{2061} & \multicolumn{1}{c|}{80.4} & \multicolumn{1}{c|}{10.8} & \multicolumn{1}{c|}{8.8} & \multicolumn{1}{c|}{4.8} & \multicolumn{1}{c|}{24.4} & \multicolumn{1}{c|}{100.0} \\ \hline
\multicolumn{1}{|l|}{vietmed\_018\_a} & \multicolumn{1}{c|}{63} & \multicolumn{1}{c|}{1527} & \multicolumn{1}{c|}{75.6} & \multicolumn{1}{c|}{12.7} & \multicolumn{1}{c|}{11.7} & \multicolumn{1}{c|}{19.6} & \multicolumn{1}{c|}{44.0} & \multicolumn{1}{c|}{100.0} \\ \hline
\multicolumn{1}{|l|}{vietmed\_018\_b} & \multicolumn{1}{c|}{192} & \multicolumn{1}{c|}{5293} & \multicolumn{1}{c|}{77.3} & \multicolumn{1}{c|}{10.0} & \multicolumn{1}{c|}{12.7} & \multicolumn{1}{c|}{6.7} & \multicolumn{1}{c|}{29.3} & \multicolumn{1}{c|}{100.0} \\ \hline
\multicolumn{1}{|l|}{vietmed\_018\_c} & \multicolumn{1}{c|}{118} & \multicolumn{1}{c|}{2761} & \multicolumn{1}{c|}{75.4} & \multicolumn{1}{c|}{12.4} & \multicolumn{1}{c|}{12.2} & \multicolumn{1}{c|}{7.4} & \multicolumn{1}{c|}{32.0} & \multicolumn{1}{c|}{100.0} \\ \hline
\multicolumn{1}{|l|}{vietmed\_018\_d} & \multicolumn{1}{c|}{20} & \multicolumn{1}{c|}{412} & \multicolumn{1}{c|}{51.7} & \multicolumn{1}{c|}{20.1} & \multicolumn{1}{c|}{28.2} & \multicolumn{1}{c|}{5.1} & \multicolumn{1}{c|}{53.4} & \multicolumn{1}{c|}{100.0} \\ \hline
\multicolumn{1}{|l|}{vietmed\_018\_e} & \multicolumn{1}{c|}{5} & \multicolumn{1}{c|}{76} & \multicolumn{1}{c|}{48.7} & \multicolumn{1}{c|}{27.6} & \multicolumn{1}{c|}{23.7} & \multicolumn{1}{c|}{5.3} & \multicolumn{1}{c|}{56.6} & \multicolumn{1}{c|}{100.0} \\ \hline
\multicolumn{1}{|l|}{vietmed\_018\_f} & \multicolumn{1}{c|}{25} & \multicolumn{1}{c|}{639} & \multicolumn{1}{c|}{64.6} & \multicolumn{1}{c|}{20.5} & \multicolumn{1}{c|}{14.9} & \multicolumn{1}{c|}{6.9} & \multicolumn{1}{c|}{42.3} & \multicolumn{1}{c|}{100.0} \\ \hline
\multicolumn{1}{|l|}{vietmed\_019\_a} & \multicolumn{1}{c|}{58} & \multicolumn{1}{c|}{1490} & \multicolumn{1}{c|}{77.4} & \multicolumn{1}{c|}{11.2} & \multicolumn{1}{c|}{11.3} & \multicolumn{1}{c|}{7.0} & \multicolumn{1}{c|}{29.6} & \multicolumn{1}{c|}{100.0} \\ \hline
\multicolumn{1}{|l|}{vietmed\_019\_b} & \multicolumn{1}{c|}{116} & \multicolumn{1}{c|}{2776} & \multicolumn{1}{c|}{78.2} & \multicolumn{1}{c|}{10.5} & \multicolumn{1}{c|}{11.3} & \multicolumn{1}{c|}{6.6} & \multicolumn{1}{c|}{28.4} & \multicolumn{1}{c|}{100.0} \\ \hline
\multicolumn{1}{|l|}{vietmed\_023} & \multicolumn{1}{c|}{390} & \multicolumn{1}{c|}{7414} & \multicolumn{1}{c|}{86.8} & \multicolumn{1}{c|}{7.7} & \multicolumn{1}{c|}{5.5} & \multicolumn{1}{c|}{4.4} & \multicolumn{1}{c|}{17.6} & \multicolumn{1}{c|}{96.7} \\ \hline
\multicolumn{1}{|l|}{vietmed\_024} & \multicolumn{1}{c|}{376} & \multicolumn{1}{c|}{7425} & \multicolumn{1}{c|}{86.9} & \multicolumn{1}{c|}{6.3} & \multicolumn{1}{c|}{6.7} & \multicolumn{1}{c|}{4.9} & \multicolumn{1}{c|}{18.0} & \multicolumn{1}{c|}{97.6} \\ \hline
\multicolumn{1}{|l|}{vietmed\_025\_a} & \multicolumn{1}{c|}{101} & \multicolumn{1}{c|}{2280} & \multicolumn{1}{c|}{82.3} & \multicolumn{1}{c|}{9.3} & \multicolumn{1}{c|}{8.4} & \multicolumn{1}{c|}{5.1} & \multicolumn{1}{c|}{22.9} & \multicolumn{1}{c|}{98.0} \\ \hline
\multicolumn{1}{|l|}{vietmed\_025\_b} & \multicolumn{1}{c|}{91} & \multicolumn{1}{c|}{1838} & \multicolumn{1}{c|}{83.2} & \multicolumn{1}{c|}{9.0} & \multicolumn{1}{c|}{7.7} & \multicolumn{1}{c|}{6.4} & \multicolumn{1}{c|}{23.1} & \multicolumn{1}{c|}{98.9} \\ \hline
\multicolumn{1}{|l|}{vietmed\_026} & \multicolumn{1}{c|}{21} & \multicolumn{1}{c|}{355} & \multicolumn{1}{c|}{56.3} & \multicolumn{1}{c|}{27.3} & \multicolumn{1}{c|}{16.3} & \multicolumn{1}{c|}{7.6} & \multicolumn{1}{c|}{51.3} & \multicolumn{1}{c|}{100.0} \\ \hline
\multicolumn{1}{|l|}{vietmed\_027\_a} & \multicolumn{1}{c|}{29} & \multicolumn{1}{c|}{710} & \multicolumn{1}{c|}{86.1} & \multicolumn{1}{c|}{6.6} & \multicolumn{1}{c|}{7.3} & \multicolumn{1}{c|}{5.8} & \multicolumn{1}{c|}{19.7} & \multicolumn{1}{c|}{100.0} \\ \hline
\multicolumn{1}{|l|}{vietmed\_027\_b} & \multicolumn{1}{c|}{64} & \multicolumn{1}{c|}{1454} & \multicolumn{1}{c|}{75.8} & \multicolumn{1}{c|}{14.9} & \multicolumn{1}{c|}{9.4} & \multicolumn{1}{c|}{6.4} & \multicolumn{1}{c|}{30.6} & \multicolumn{1}{c|}{100.0} \\ \hline
\multicolumn{1}{|l|}{vietmed\_028\_a} & \multicolumn{1}{c|}{106} & \multicolumn{1}{c|}{2617} & \multicolumn{1}{c|}{83.5} & \multicolumn{1}{c|}{8.7} & \multicolumn{1}{c|}{7.9} & \multicolumn{1}{c|}{4.6} & \multicolumn{1}{c|}{21.1} & \multicolumn{1}{c|}{100.0} \\ \hline
\multicolumn{1}{|l|}{vietmed\_028\_b} & \multicolumn{1}{c|}{21} & \multicolumn{1}{c|}{475} & \multicolumn{1}{c|}{76.4} & \multicolumn{1}{c|}{14.5} & \multicolumn{1}{c|}{9.1} & \multicolumn{1}{c|}{6.5} & \multicolumn{1}{c|}{30.1} & \multicolumn{1}{c|}{95.2} \\ \hline
\multicolumn{1}{|l|}{vietmed\_029} & \multicolumn{1}{c|}{92} & \multicolumn{1}{c|}{2240} & \multicolumn{1}{c|}{84.6} & \multicolumn{1}{c|}{7.7} & \multicolumn{1}{c|}{7.7} & \multicolumn{1}{c|}{5.7} & \multicolumn{1}{c|}{21.1} & \multicolumn{1}{c|}{100.0} \\ \hline
\multicolumn{1}{|l|}{\textbf{Sum/Avg}} & \multicolumn{1}{c|}{3437} & \multicolumn{1}{c|}{76136} & \multicolumn{1}{c|}{76.5} & \multicolumn{1}{c|}{13.1} & \multicolumn{1}{c|}{10.3} & \multicolumn{1}{c|}{5.5} & \multicolumn{1}{c|}{\textbf{29.0}} & \multicolumn{1}{c|}{98.9} \\ \hline
\multicolumn{1}{|l|}{\textbf{Mean}} & \multicolumn{1}{c|}{127.3} & \multicolumn{1}{c|}{2819.9} & \multicolumn{1}{c|}{75.5} & \multicolumn{1}{c|}{13.0} & \multicolumn{1}{c|}{11.5} & \multicolumn{1}{c|}{6.1} & \multicolumn{1}{c|}{30.6} & \multicolumn{1}{c|}{99.3} \\ \hline
\multicolumn{1}{|l|}{\textbf{S.D.}} & \multicolumn{1}{c|}{129.6} & \multicolumn{1}{c|}{2743.3} & \multicolumn{1}{c|}{11.3} & \multicolumn{1}{c|}{7.2} & \multicolumn{1}{c|}{5.1} & \multicolumn{1}{c|}{2.9} & \multicolumn{1}{c|}{11.9} & \multicolumn{1}{c|}{1.3} \\ \hline
\multicolumn{1}{|l|}{\textbf{Median}} & \multicolumn{1}{c|}{86.0} & \multicolumn{1}{c|}{1838.0} & \multicolumn{1}{c|}{77.6} & \multicolumn{1}{c|}{11.2} & \multicolumn{1}{c|}{9.7} & \multicolumn{1}{c|}{5.7} & \multicolumn{1}{c|}{28.4} & \multicolumn{1}{c|}{100.0} \\ \hline
\caption{Breakdown per speaker on the Vietnamese test set of the Hybrid ASR results in Table \ref{tab.AED_vs_Hybrid}. Two pre-trained wav2vec 2.0 models were used for fine-tuning on the Vietnamese set: XLSR-53-Viet and w2v2-Viet, leading to WERs on test set  28.8\%, 29.0\% respectively.\\
Column from left to right is: Speaker ID, Number of sentences, Number of words, Corrections, Substitution Errors, Deletion Errors, Insertion Errors, Word-Error-Rate, Sentence-Error-Rate.}
\label{tab.breakdown_HybridASR_perSpeaker}
\end{longtable}
\twocolumn

\section{Full Error Analysis}
\label{sec:error_analysis_linguistic_perspective}
Figure \ref{ASR_errors_example} shows an example of common ASR errors from the ASR output compared to the corresponding ground truth transcript. Three ASR errors considered are substitutions, deletions, and insertions.

Below is the full error analysis based on the linguistic perspective for all 5 languages: English, Vietnamese, Chinese, French, and German.

\subsection{English}
Our error analysis of the ASR system revealed several phonological issues that affected the performance of the model. One significant issue involves the minimal phonological distance between certain vowel sounds, particularly in minimal pairs such as "long" vs. "lung" and "pen" vs. "pan". Due to the close proximity of these sounds in the phonetic space, the model often confuses them, leading to clinically significant errors, such as transcribing "lung cancer" as "long cancer".

Another source of error is related to the use of weak forms in speech, where certain words are pronounced in a reduced or less distinct manner. This results in frequent misrecognitions, such as interpreting "our" as "are", "and" as "in", "for" as "very", and even more complex substitutions like "earlierologist" for the phrase "earlier I was just". Additionally, numerical errors are common; for instance, the model may interpret "4 to 5" as "45", which could lead to critical inaccuracies in medical records. This type of substitution also extends to domain-specific terminology, such as transcribing "system that" as "systemic".

In addition, discrepancies were identified at the beginning and end of the transcriptions. This issue is largely attributed to inconsistencies between the training and testing conditions: the dataset was annotated using long-form audio segments, yet the model was trained and evaluated with short-form audio inputs. This mismatch creates boundary errors and negatively affects the model’s ability to capture context, leading to truncation or overlap in predictions. In particular, this problem is not limited to English but has also been observed in other languages, indicating a systematic problem in handling different input formats during ASR processing.

\subsection{Vietnamese}
In the Vietnamese test set, a detailed analysis of ASR errors reveals that several phonological characteristics of the Vietnamese language pose significant challenges for model performance. Vietnamese is a tonal language with a complex phonetic system that includes a variety of vowels, consonants, and six distinct tones, all of which carry meaning and are integral to word differentiation \cite{horn2004vietnamese}. As a result, the ASR system often struggles with minimal phonetic contrasts, particularly when dealing with similar-sounding phonemes and tones.

\textbf{Vowel confusion}: Vietnamese vowels exhibit subtle distinctions, especially in terms of vowel height and backness. Pairs such as "cái" vs. "cứ" demonstrate this challenge. The model frequently confuses these due to their similar articulatory features and acoustic proximity. For instance, "cái" (meaning "thing" or "classifier for objects") and "cứ" (meaning "to keep doing something") differ primarily in vowel quality, but the ASR system often fails to capture this distinction, leading to misrecognition.

\textbf{Consonant ambiguity}: Consonant sounds in Vietnamese can also present difficulties, particularly when the phonemes are produced with similar places of articulation. An example is "nó" (he/she/it) vs. "nói" (to speak), where the confusion arises due to the similarity in nasal sounds and the rapid articulation of connected speech. Similarly, the pair "bác" (uncle/aunt) and "mắc" (to catch/to be caught) are often misrecognized due to the shared stop consonant sounds, complicated further by the presence of nasal or plosive release.

\textbf{Tonal ambiguity}: Vietnamese tones are particularly problematic for ASR systems, as they are both lexically and syntactically significant. The six tones in Vietnamese include level, rising, falling, broken, creaky, and low tones, which can completely change the meaning of a word. For instance, the pair "năng ngọng" (meaning "slurred speech") and "nặng nhọc" (meaning "laborious") illustrates how the model struggles to distinguish between tones, leading to semantically incorrect transcriptions. The difference between these phrases lies in tone distinctions, which are subtle and can be easily confounded by background noise or speaker variability.

\textbf{Gender and regional variations}: Furthermore, phonological variability due to gender differences (for example, male vs. female voice pitch) and regional dialects (Northern, Central, and Southern accents) further complicates the ability of the ASR system to correctly distinguish words of similar sound. For example, "nử" (variant pronunciation for some speakers, typically Northern) and "nữ" (female) differ mainly in tone and vowel length, which may be pronounced differently across dialects, increasing the error rate.

These types of phonological errors highlight the need for enhanced acoustic modeling that can account for the intricate vowel and consonant distinctions and the tonal nature of Vietnamese, especially in the medical domain. It also underscores the importance of incorporating a diverse set of training data that reflect different regional accents and speech patterns for patients and doctors to improve the robustness of the medical ASR system in Vietnamese language contexts.

\subsection{Chinese}
\begin{CJK}{UTF8}{gbsn}
A primary source of errors arises from minimal pairs that differ solely in tonal pronunciation or involve homophones, both of which are highly prevalent in Mandarin Chinese. Given that Mandarin is a tonal language with four distinct tones (plus a neutral tone), words that share similar phonetic sounds but differ in tone can easily be confused by ASR systems \cite{jongman2006perception}. This tonal ambiguity leads to significant transcription errors, especially in medical contexts where precision is crucial.

For instance in our test set, words like 麻闭 (mábì) and 麻痹 (má bì), or 跟本 (gēnběn) and 根本 (gēn běn), demonstrate how tonal distinctions are critical for differentiating between distinct meanings. Similarly, homophones such as 以 (yǐ) and 已 (yǐ), or 是 (shì) and 适 (shì), present further challenges, as the ASR system struggles to disambiguate words with identical phonetic pronunciation but different meanings. The error is compounded by the context-dependent nature of these terms, which requires a sophisticated understanding of the surrounding text to accurately differentiate them.

Additionally, errors are frequently caused by words that sound alike but differ in their meaning, as seen in examples like 代 (dài) vs 待 (dài) or 其二 (qí èr) vs 妻儿 (qī ér). In the medical domain, such mistakes can lead to severe clinical misinterpretations, affecting patient safety. For example, confusion between 没 (méi) (not) and 霉 (méi) (mold) could result in significant differences in the interpretation of a patient's condition or diagnosis.

Another frequent source of error is phonetic approximation in the sound space, where slight variations in pronunciation result in incorrect word predictions. Examples include 到路 (dào lù) vs 倒漏 (dào lòu) and 一确的 (yī què de) vs 一切都 (yīqiè dōu). These phonetic approximations arise due to the ASR system's inability to distinguish subtle differences, particularly in connected speech where articulation may be less clear. Such approximations can be particularly problematic in medical transcription, where terms like 答案 (dá'àn) (answer) being mistaken for 大碍 (dà ài) (serious problem) could alter the intended meaning of a clinical statement.
\end{CJK}

\subsection{French}
The errors encountered in the medical domain's ASR systems can be attributed to various phonological challenges, especially in datasets with languages like French, where the close proximity of certain phonemes in the acoustic space leads to frequent misinterpretations. These challenges typically arise from the inherent acoustic similarity between phonemes or word pairs that sound alike but have different meanings or spelling, often referred to as homophones or near-homophones For instance, in the French language, there are numerous vowel and consonant pairs that share similar acoustic characteristics but differ in meaning, making them susceptible to confusion. Some notable examples include:

\begin{itemize}
    \item "attention" vs "ah tiens": Both phrases have similar phonetic structures, but the former is a common French word meaning "careful" or "attention", while the latter is a colloquial expression that might refer to a surprise or exclamation. A misinterpretation of these terms could lead to clinical miscommunication in situations requiring urgency or specific instructions.
    \item "engardré" vs "encadré": These words differ by a single vowel sound, but the first ("engardré") is a non-standard form or a potential misheard word, while the latter ("encadré") means "framed" in French. Such phonetic ambiguity can easily result in incorrect transcription, especially when the ASR model is unable to distinguish between similar-sounding terms within the context of a medical discussion.
    \item "à mettre" vs "est maître": The phrase "à mettre" (meaning "to put") is often misheard as "est maître" (meaning "is the master"), as both phrases have a similar rhythm and vowel-consonant structure. In medical settings, such confusion could mislead the interpretation of a patient's condition or instructions for care.
    \item "bonchique" vs "bronchite": A typical error arises when the ASR system confuses "bronchite" (bronchitis) with a distorted form like "bonchique". This could be catastrophic in medical contexts, as bronchitis refers to a serious respiratory condition, and an error here could delay proper diagnosis or treatment.
    \item "choléraux" vs "cholestérol": The acoustic similarity between "choléraux" (a non-standard or incorrect form) and "cholestérol" (cholesterol) presents another challenge. Cholesterol is a critical term in medical diagnostics, and errors in its transcription could result in the omission of vital health information, leading to inaccurate clinical assessments or interventions.
    \item "mé" vs "mais": The confusion between "mé" (which can be a shorthand or mispronunciation of "mais" meaning "but") is another example. Such errors are especially significant in medical contexts where subtle linguistic distinctions, even in less formal speech, can alter the meaning of a diagnosis or treatment plan.
\end{itemize}

\subsection{German}
In the context of ASR error analysis within the medical domain, our German test set presents distinct challenges that stem from both phonological and orthographic factors, which significantly affect the model’s accuracy and performance.

Firstly, the issue of phonological proximity is particularly noticeable in minimal pairs—pairs of words that differ only in one sound. In the German language, small phonological differences between vowels in minimal pairs can cause considerable confusion for ASR models, as these systems often struggle to accurately distinguish between such similar-sounding words. For instance, the words "verschmerzen" (to suffer pain) and "vor Schmerzen" (before pain) have a very slight phonetic distinction, yet they represent entirely different meanings, potentially leading to misinterpretation by the ASR system. Similarly, words like "anestätiger" (anesthetist) and "Lokalanästhetikasalbe" (local anesthetic cream) contain subtle phonetic differences that can cause errors in transcription, especially when such words are transcribed without appropriate context or clarity.

Secondly, the orthographic characteristics of the German language further complicate ASR performance. German has a system of capitalization where nouns and imperative verbs are capitalized, while adjectives, adverbs, verbs, and other parts of speech are written in lowercase. This capitalization rule is not just a grammatical convention, but a semantic one, as it helps distinguish between different parts of speech and the meaning of the sentence. ASR models that fail to accurately capture these distinctions often produce errors that are both semantically and syntactically problematic. For example, "venenzugang" (venous access) vs "Venenzugang" (with proper capitalization) may lead to a loss of meaning or context in the transcribed text. Similarly, confusion between "komme" (come) and "Komme" (I come, in the imperative) can alter the intended message, especially in medical contexts where the clarity of instructions is critical.

\begin{table*}[hbt!]
\centering  
\small
\begin{tabular}{cp{2cm}p{10cm}  }
\toprule
\multicolumn{3}{l}{\bf Example}  \\
\midrule
\multirow{3}{*}{English}
& ASR output & \textcolor{red!70!black}{\textbf{sea}} you don't really see any \textcolor{red!70!black}{\textbf{affect}} the brown \textcolor{red!70!black}{\textbf{apocalyse}} tissue activity, but at the high \textcolor{red!70!black}{\textbf{BMW}}, now, you will start to see a \textcolor{green!70!black}{\textbf{uh}} \textcolor{green!70!black}{\textbf{uhm}} protective effect where those individuals had lower \textcolor{red!70!black}{\textbf{glyceryl}}.
\\
& Ground truth & \textcolor{blue!70!black}{\textbf{only}} see you don't really see any effect \textcolor{blue!70!black}{\textbf{of}} the brown adipose tissue activity, but at the high BMI, now, you will start to see a protective effect where those individuals had lower glycemia.\\

\multirow{3}{*}{Chinese}
& ASR output & \begin{CJK}{UTF8}{gbsn}们新安装的那\textcolor{red!70!black}{\textbf{更新}}门是在这里， 然后我们看一\textcolor{green!70!black}{\textbf{个}}下有没有倒漏的问题， 有没有狭窄的那个情况。\end{CJK}
\\
& Ground truth & \begin{CJK}{UTF8}{gbsn}\textcolor{blue!70!black}{\textbf{我}}们新安装的那个心门是在这里， 然后我们看一下有没有倒漏的问题， 有没有狭窄的那个情况。\end{CJK}\\

\multirow{3}{*}{French}
& ASR output & arrivez à \textcolor{green!70!black}{\textbf{à}} sortir un peu ou pas du tout 36 \textcolor{red!70!black}{\textbf{tempérament}} c'est bien vous savez vous avez un mix entre la \textcolor{red!70!black}{\textbf{broncoid}} l'insuffisance cardiaque et tout ce qui.
\\
& Ground truth & arrivez à sortir un peu ou pas du tout 36 \textcolor{blue!70!black}{\textbf{la}} température c'est bien vous savez vous avez un mix entre la bronchite l'insuffisance cardiaque et tout ce qui\\

\multirow{3}{*}{German}
& ASR output & Haben Sie Allergiepass oder einen \textcolor{red!70!black}{\textbf{Reisepass}}? Dann könnte ich da mal nachschauen, ob \textcolor{green!70!black}{\textbf{mal}} ein spezielles \textcolor{red!70!black}{\textbf{Antibiotikern}} eingetragen worden ist. ich habe beides, da \textcolor{green!70!black}{\textbf{ja}} steht alles drin. Die bringt mein
\\
& Ground truth & Haben Sie \textcolor{blue!70!black}{\textbf{einen}} Allergiepass oder einen Patientenpass? Dann könnte ich da mal nachschauen, ob ein spezielles Antibiotikum eingetragen worden ist. \textcolor{blue!70!black}{\textbf{Ja}}, ich habe beides, da steht alles drin. Die bringt mein\\

\multirow{3}{*}{Vietnamese}
& ASR output & bản thân và \textcolor{green!70!black}{\textbf{ừ}} rộng hơn là \textcolor{green!70!black}{\textbf{là}} vì sức khỏe cộng đồng thưa quý \textcolor{red!70!black}{\textbf{dị}} tại việt nam nguyên tắc huyết khối \textcolor{red!70!black}{\textbf{tiễn}} mạch bệnh \textcolor{red!70!black}{\textbf{mặt}} máu
\\
& Ground truth & bản thân và rộng hơn là vì sức khỏe cộng đồng thưa quý vị tại việt nam nguyên tắc huyết khối tĩnh mạch \textcolor{blue!70!black}{\textbf{là}} bệnh mạch máu\\

\bottomrule
\end{tabular}
\caption{An example of ASR errors from ASR output (top) compared to the corresponding ground truth transcript (bottom). Errors are annotated as: substitutions in \textcolor{red!70!black}{\textbf{red}},
deletions in \textcolor{blue!70!black}{\textbf{blue}}, and insertions in \textcolor{green!70!black}{\textbf{green}}.
}
\label{ASR_errors_example}
\end{table*}
\end{document}